\documentclass{bmvc2k}

\usepackage{amssymb}
\usepackage{bm}
\usepackage{capt-of}
\usepackage{hyperref}
\usepackage{multirow}
\newcommand{\bhline}[1]{\noalign{\hrule height #1}}

\makeatletter
\newcommand{\figcaption}[1]{\def\@captype{figure}\caption{#1}}
\newcommand{\tblcaption}[1]{\def\@captype{table}\caption{#1}}
\makeatother

\graphicspath{{figs/}}

\usepackage{xr}
\title{Class-Distinct and Class-Mutual Image Generation with GANs}

\addauthor{Takuhiro Kaneko}{t.kaneko@mi.t.u-tokyo.ac.jp}{1}
\addauthor{Yoshitaka Ushiku}{ushiku@mi.t.u-tokyo.ac.jp}{1}
\addauthor{Tatsuya Harada$^{\text{1,}}$}{harada@mi.t.u-tokyo.ac.jp}{2}

\addinstitution{
  Graduation School of\\
  Information Science and Technology\\
  The University of Tokyo\\
  Tokyo, Japan
}
\addinstitution{
  RIKEN Center for\\
  Advanced Intelligence Project\\
  Tokyo, Japan
}

\runninghead{T. Kaneko \etal}{Class-Distinct and Class-Mutual Image Generation}

\def\eg{\emph{e.g}\bmvaOneDot}

\def\etal{\emph{et al}\bmvaOneDot}
\def\ie{\emph{i.e}\bmvaOneDot}

\begin{document}

\maketitle

\begin{abstract}
  Class-conditional extensions of generative adversarial networks (GANs), such as auxiliary classifier GAN (AC-GAN) and conditional GAN (cGAN), have garnered attention owing to their ability to decompose representations into class labels and other factors and to boost the training stability. However, a limitation is that they assume that each class is separable and ignore the relationship between classes even though class overlapping frequently occurs in a real-world scenario when data are collected on the basis of diverse or ambiguous criteria. To overcome this limitation, we address a novel problem called class-distinct and class-mutual image generation, in which the goal is to construct a generator that can capture between-class relationships and generate an image selectively conditioned on the class specificity. To solve this problem without additional supervision, we propose classifier's posterior GAN (CP-GAN), in which we redesign the generator input and the objective function of AC-GAN for class-overlapping data. Precisely, we incorporate the classifier's posterior into the generator input and optimize the generator so that the classifier's posterior of generated data corresponds with that of real data. We demonstrate the effectiveness of CP-GAN using both controlled and real-world class-overlapping data with a model configuration analysis and comparative study. Our code is available at \url{https://github.com/takuhirok/CP-GAN/}.
\end{abstract}

\section{Introduction}
\label{sec:intro}

\begin{figure*}[t]
  \centering
  \includegraphics[width=0.97\textwidth]{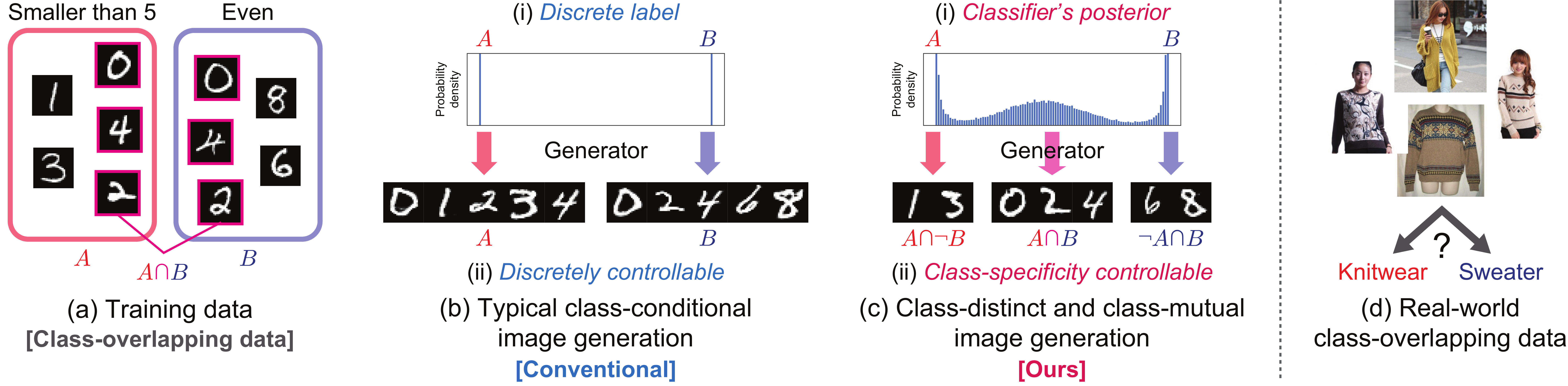}
  \vspace{-2mm}
  \caption{
    Example of class-distinct and class-mutual image generation. Given class-overlapping data (a), a typical class-conditional image generation model (\eg, AC-GAN; (b)) is optimized conditioned on discrete labels (b-i) and generates data of each class separately (b-ii). In contrast, our class-distinct and class-mutual image generation model (\ie, CP-GAN; (c)) represents between-class relationships in the generator input using the classifier's posterior (c-i) and generates an image conditioned on the class specificity (c-ii). Classes frequently overlap in a real-world scenario when classes are confusing (d).}  
  \label{fig:example}
  \vspace{-3mm}
\end{figure*}

In computer vision and machine learning, generative adversarial networks (GANs)~\cite{IGoodfellowNIPS2014} have become a prominent model owing to their ability to represent high-dimensional data in a compact latent space and to generate high-fidelity data~\cite{TKarrasICLR2018,TMiyatoICLR2018b,HZhangICML2019,ABrockICLR2019,TKarrasCVPR2019}. In particular, class-conditional extensions of GANs (\eg, auxiliary classifier GAN (AC-GAN)~\cite{AOdenaICML2017} and conditional GAN (cGAN)~\cite{MMirzaArXiv2014,TMiyatoICLR2018}) have garnered attention owing to their two strong properties: (1) The supervision of class labels makes it possible to decompose representations into class labels and other factors. This allows the generator to generate an image selectively conditioned on class labels~\cite{MMirzaArXiv2014,AOdenaICML2017,TKanekoCVPR2017,ZZhangCVPR2017,TMiyatoICLR2018,TKanekoCVPR2018}. (2) The additional information simplifies the learned target from an overall distribution to the class-conditional distributions. This contributes to stabilize the GAN training~\cite{AOdenaICML2017,TMiyatoICLR2018,HZhangICML2019,ABrockICLR2019}.

However, a limitation is that they assume that each class is separable and ignore the relationship between classes. This not only restricts applications but also causes difficulty in modeling class-overlapping data even though class overlapping frequently occurs in a real-world scenario when data are collected on the basis of diverse or ambiguous criteria. Figure~\ref{fig:example} shows an example. In Figure~\ref{fig:example}(a), class $A$ includes digits that are smaller than five and class $B$ contains even digits. In this case, ``0,'' ``2,'' and ``4'' belong to both classes. Given such \textit{class-overlapping} data, a typical class-conditional image generation model (\eg, AC-GAN~\cite{AOdenaICML2017}) is optimized conditioned on \textit{discrete} labels (Figure~\ref{fig:example}(b-i)) and generates data of each class \textit{separately} (Figure~\ref{fig:example}(b-ii)) without considering the between-class relationships.

To remedy this drawback, we address a novel problem called \textit{class-distinct and class-mutual image generation}, in which the goal is to construct a generator that can capture between-class relationships and generate an image selectively on the basis of the \textit{class specificity}. To solve this problem without additional supervision, we propose \textit{classifier's posterior GAN (CP-GAN)}, in which we redesign the generator input and the objective function of AC-GAN for \textit{class-overlapping} data. Precisely, we employ the classifier's posterior to represent the between-class relationships and incorporate it into the generator input, as shown in Figure~\ref{fig:example}(c-i). Additionally, we optimize the generator so that the classifier's posterior of generated data corresponds with that of real data. This formulation allows CP-GAN to capture the between-class relationships in a data-driven manner and to generate an image conditioned on the \textit{class specificity}, as shown in Figure~\ref{fig:example}(c-ii).

To the best of our knowledge, class-distinct and class-mutual image generation is a novel task not sufficiently examined in previous studies. To advance this research, in the experiments, we introduce controlled class-overlapping data on CIFAR-10~\cite{AKrizhevskyTech2009} and report the benchmark performance with a model configuration analysis and comparative study. Classes frequently overlap in a real-world scenario when classes are confusing, as shown in Figure~\ref{fig:example}(d). Hence, we also evaluated CP-GAN using such real-world class-overlapping data.

Overall, our contributions are summarized as follows:
\begin{itemize}
  \vspace{-2mm}
  \setlength{\parskip}{1pt}
  \setlength{\itemsep}{1pt}
\item We tackle a novel problem called \textit{class-distinct and class-mutual image generation}: given \textit{class-overlapping} data, the goal is to construct a generator that can capture between-class relationships and generate an image selectively conditioned on the \textit{class specificity}.
\item To solve this problem, we propose \textit{CP-GAN}, in which we redesign the generator input and the objective function of AC-GAN for \textit{class-overlapping} data. This formulation allows CP-GAN to solve this problem without additional supervision.
\item We demonstrate the effectiveness of CP-GAN using both controlled and real-world class-overlapping data with a model configuration analysis and comparative study.
\item As further analysis, we also demonstrate the generality of CP-GAN by applying it to image-to-image translation. See Appendix~\ref{sec:exp_pix2pix} for details.
\item Our code is available at \url{https://github.com/takuhirok/CP-GAN/}.
  \vspace{-2mm}
\end{itemize}

\section{Related work}
\label{sec:related}

\textbf{Deep generative models.}
Along with GANs, prominent three deep generative models are variational autoencoders (VAEs)~\cite{DKingmaICLR2014,DRezendeICML2014}, autoregressive models (ARs)~\cite{AOordICML2016}, and flow-based models (Flows)~\cite{LDinhICLR2017}. All these models have pros and cons. A well-known weakness of GANs is training instability; however, this has been improved in recent work~\cite{EDentonNIPS2015,ARadfordICLR2016,TSalimansNIPS2016,JZhaoICLR2017,MArjovskyICLR2017,MArjovskyICML2017,XMaoICCV2017,IGulrajaniNIPS2017,TKarrasICLR2018,XWeiICLR2018,TMiyatoICLR2018b,LMeschederICML2018,HZhangICML2019,ABrockICLR2019,TKarrasCVPR2019}. In this study, we focus on GANs owing to their flexibility in designing the latent space. This makes it easy to incorporate conditional information, such as our classifier's posterior. Conditional extensions have been proposed for other models~\cite{DKingmaNIPS2014,XYanECCV2016,EMansimovICLR2016,AOordNIPS2016,SReedICLRW2017}, and incorporating our ideas into them is a promising direction of future work.

\smallskip\noindent
\textbf{Disentangled representation learning.}
Naive GANs do not have an explicit structure in the latent space. Hence, their latent variables may be used by the generator in a highly entangled manner. To solve this problem, recent studies have incorporated supervision into the networks~\cite{MMirzaArXiv2014,AOdenaICML2017,LTranCVPR2017,TKanekoCVPR2017,TMiyatoICLR2018,TKanekoCVPR2018,SReedICML2016,HZhangICCV2017,HZhangArXiv2017,XTaoCVPR2018,SReedNIPS2016,TKanekoCVPR2019}. However, their learnable representations are restricted to the given supervision. To overcome this limitation, unsupervised~\cite{XChenNIPS2016,TKanekoCVPR2018} and weakly supervised~\cite{AMakhzaniNIPS2016,MMathieuNIPS2016,TKanekoCVPR2017,TKanekoCVPR2018} models have been also proposed. Our proposed CP-GAN is a weakly supervised model because it learns a \textit{class-specificity} controllable model only using weak supervision (\ie, \textit{discrete} labels). The difference from the previous models is that CP-GAN incorporates \textit{class specificity}. In this aspect, class-distinct and class-mutual image generation is different from typical \textit{class-wise interpolation} (or category morphing in \cite{TMiyatoICLR2018}) because the latter interpolates between classes regardless of class specificity. We empirically demonstrate this difference in Figure~\ref{fig:cifar10_interp} in the Appendix. Additionally, We further discuss the relationships between CP-GAN and previous class-conditional extensions of GANs in Section~\ref{subsec:relationship}.

\section{Class-distinct and class-mutual image generation}
\label{sec:method}

\subsection{Notations and problem statement}
\label{subsec:notation}

We first define notations and the problem statement. We use superscripts $r$ and $g$ to denote the real distribution and the generative distribution, respectively. Let ${\bm x} \in {\cal X}$ and ${\bm y} \in {\cal Y}$ be the image and the corresponding class label, respectively. Here, ${\cal X}$ is the image space ${\cal X} \subseteq \mathbb{R}^d$, where $d$ is the dimension of the image. ${\cal Y}$ is the label space ${\cal Y} = \{ {\bm y}: {\bm y} \in \{ 0, 1 \}^c, {\bm 1}^{\top} {\bm y} = 1 \}$, where $c$ is the number of classes and we use a one-hot vector form to represent labels. We address the \textit{instance-dependent class overlapping}: class overlapping that occurs depending on the image content. In such a situation, \textit{class-distinct} and \textit{class-mutual} states are determined relying on the underlying class posterior $p^r({\bm y} | {\bm x})$. For example, when $p^r({\bm y} | {\bm x})$ is close to 1 for a specific class, ${\bm x}$ is a \textit{class-distinct} image. In contrast, when $p^r({\bm y} | {\bm x})$ has some values for multiple classes, ${\bm x}$ is a \textit{class-mutual} image.

\begin{figure*}[t]
  \centering
  \includegraphics[width=0.99\textwidth]{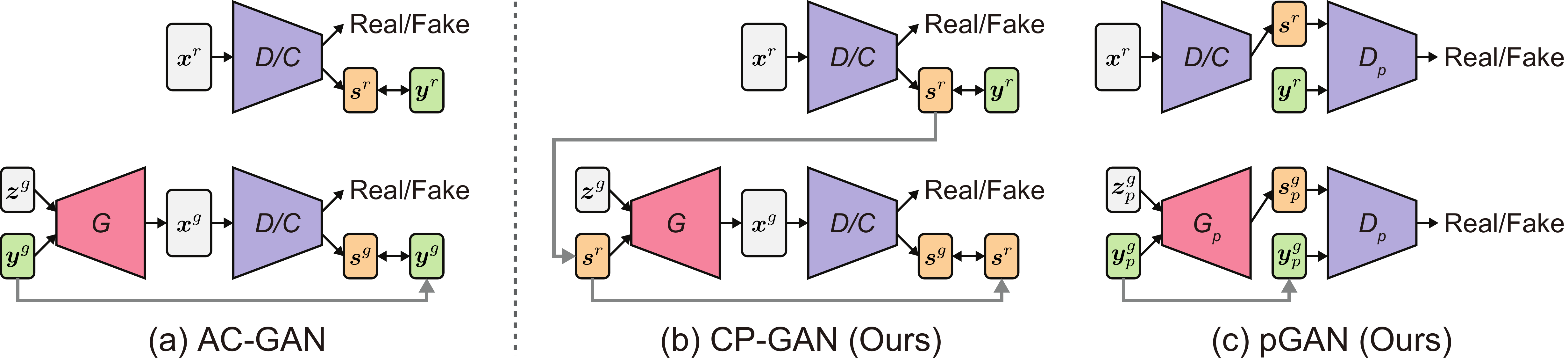}
  \vspace{-2mm}
  \caption{Comparison of AC-GAN (a) and CP-GAN (b). A green rectangle indicates a discrete label (or hard label), while an orange rectangle indicates a classifier's posterior (or soft label). In CP-GAN, we redesign the generator input and the objective function of AC-GAN to construct a generator that is conditioned on the class specificity. CP-GAN requires a classifier's posterior for real data (${\bm s}^r$) for generating an image. To mitigate this requirement, we introduce pGAN (c) that learns a classifier's posterior distribution.}  
  \label{fig:model}
  \vspace{-2mm}
\end{figure*}

Our goal is, given \textit{discretely} labeled data $({\bm x}^r, {\bm y}^r) \sim p^r({\bm x}, {\bm y})$, to construct a class-distinct and class-mutual image generator that can selectively generate an image conditioned on the \textit{class specificity} (\ie, $p^r({\bm y} | {\bm x})$). This is challenging for typical conditional GANs because they learn a generator conditioned on the observable \textit{discrete} ${\bm y}^r$ and cannot incorporate the \textit{class specificity}. To solve this problem without additional supervision, we develop CP-GAN, in which we redesign the generator input and the objective function of AC-GAN, as shown in Figure~\ref{fig:model}(b). In the next sections, we review AC-GAN (which is the baseline of CP-GAN), clarify the limitations of AC-GAN in class-overlapping data, and introduce CP-GAN. Finally, we summarize the relationships with previous class-conditional extensions of GANs.

\subsection{Baseline: AC-GAN}
\label{subsec:acgan}

AC-GAN~\cite{AOdenaICML2017} is a class-conditional extension of GAN. Its aim is to learn a conditional generator $G$ that transforms the noise ${\bm z}^g$ and the label ${\bm y}^g$ into the image ${\bm x}^g$, \ie, ${\bm x}^g = G({\bm z}^g, {\bm y}^g)$, where ${\bm z}^g$ is the noise sampled from a normal distribution $p^g({\bm z}) = {\cal N}(0, I)$ and ${\bm y}^g$ is the class label sampled from a categorical distribution $p^g({\bm y}) = \text{Cat}(K=c, p=1/c)$. To achieve this aim, AC-GAN uses two losses, namely an adversarial loss~\cite{IGoodfellowNIPS2014} and auxiliary classifier (AC) loss~\cite{AOdenaICML2017}. We illustrate the AC-GAN architecture in Figure~\ref{fig:model}(a).

\smallskip\noindent
\textbf{Adversarial loss.}
The adversarial loss is defined as
\begin{flalign}  
  \label{eqn:adv}
  {\cal L}_{\text{GAN}} = \mathbb{E}_{{\bm x}^r \sim p^r({\bm x})} [ \log D({\bm x}^r) ] + \mathbb{E}_{{\bm z}^g \sim p^g({\bm z}), {\bm y}^g \sim p^g({\bm y})} [ \log (1 - D(G({\bm z}^g, {\bm y}^g))) ],
\end{flalign}
where $D$ is a discriminator that attempts to find the best decision boundary between the real images ${\bm x}^r$ and the generated image ${\bm x}^g = G({\bm z}^g, {\bm y}^g)$ by maximizing this loss, while $G$ attempts to generate an image ${\bm x}^g$ indistinguishable by $D$ by minimizing this loss.

\smallskip\noindent
\textbf{AC loss.}
The AC losses of real images and generated images are respectively defined as
\begin{flalign}
  \label{eqn:cls_r}
  {\cal L}_{\text{AC}}^r & = \mathbb{E}_{({\bm x}^r, {\bm y}^r) \sim p^r({\bm x}, {\bm y})} [ -\log C({\bm y} = {\bm y}^r | {\bm x}^r) ] = \mathbb{E}_{({\bm x}^r, {\bm y}^r) \sim p^r({\bm x}, {\bm y})} [ -{{\bm y}^r}^{\top} \log {\bm s}^r ],
  \\
  \label{eqn:cls_g}
  {\cal L}_{\text{AC}}^g & = \mathbb{E}_{{\bm z}^g \sim p^g({\bm z}), {\bm y}^g \sim p^g({\bm y})} [ -\log C({\bm y} = {\bm y}^g | G({\bm z}^g, {\bm y}^g)) ] = \mathbb{E}_{{\bm z}^g \sim p^g({\bm z}), {\bm y}^g \sim p^g({\bm y})} [ - {{\bm y}^g}^{\top} \log {\bm s}^g ],
\end{flalign}
where ${\bm s}^r = C({\bm y} | {\bm x}^r)$ and ${\bm s}^g = C({\bm y} | G({\bm z}^g, {\bm y}^g))$ are classifier's posteriors (or soft labels) for a real image and generated image, respectively. By minimizing ${\cal L}_{\text{AC}}^r$, $C$ learns to classify the real image ${\bm x}^r$ to the corresponding class ${\bm y}^r$. Subsequently, $G$ attempts to generate an image ${\bm x}^g$ classified to the corresponding class ${\bm y}^g$, by minimizing ${\cal L}_{\text{AC}}^g$.

\smallskip\noindent
\textbf{Full objective.}
In practice, shared networks between $D$ and $C$ are commonly used~\cite{AOdenaICML2017,IGulrajaniNIPS2017}. In this configuration, the full objective is written as
\begin{flalign}
  \label{eqn:full}
  {\cal L}_{D\text{/}C} = - {\cal L}_{\text{GAN}} + \lambda^r {\cal L}_{\text{AC}}^r,
  \:\:\:\:\:
  {\cal L}_{G} = {\cal L}_{\text{GAN}} + \lambda^g {\cal L}_{\text{AC}}^g,
\end{flalign}
where $\lambda^r$ and $\lambda^g$ are trade-off parameters between the adversarial loss and the AC loss for the real image and generated image, respectively. $D$/$C$ and $G$ are optimized by minimizing ${\cal L}_{D\text{/}C}$ and ${\cal L}_G$, respectively.

\begin{figure*}[t]
  \centering
  \includegraphics[width=0.99\textwidth]{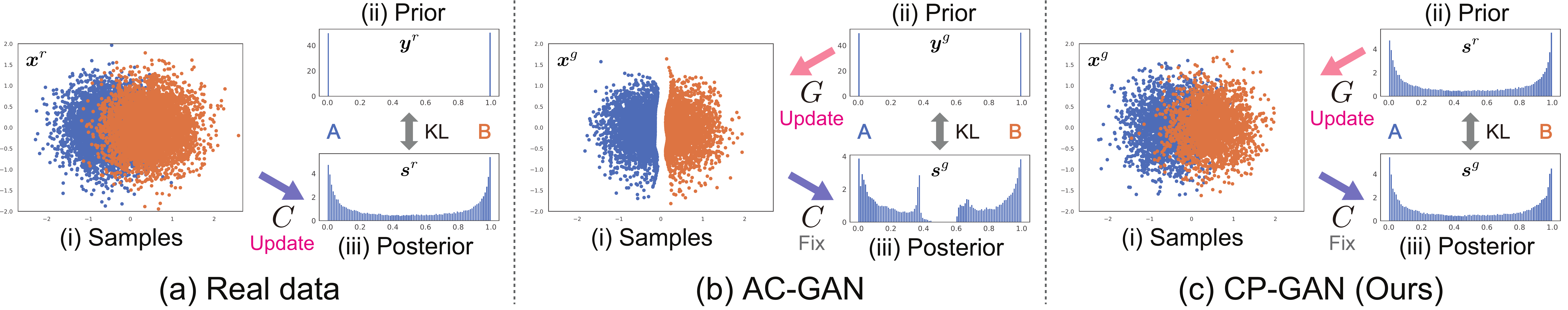}
  \vspace{-2mm}
  \caption{Comparison of training procedures in the AC losses. (a) In the AC loss of real data, $C$ is optimized to capture the class overlapping state in ${\bm x}^r$ (a-i). (b) In AC-GAN, $G$ is optimized so that the classifier's posterior for the generated data (${\bm s}^g$; b-iii) is close to the discrete prior (${\bm y}^g$; b-ii). This enforces $G$ to generate class-separate data (b-i). (c) In contrast, in CP-GAN, $G$ is optimized so that the classifier's posterior for the generated data (${\bm s}^g$; c-iii) is close to the classifier's posterior for the real data (${\bm s}^r$; c-ii). This allows $G$ to generate class-overlapping data (c-i) that are close to real data (a-i).}
  \label{fig:syn2d}
  \vspace{-2mm}
\end{figure*}

\subsection{Limitations of AC-GAN in class-overlapping data}
\label{subsec:theory}

To explain the limitations of AC-GAN in class-overlapping data, we use a toy example: data that consist of two-class Gaussian distributions with class overlapping, as shown in Figure~\ref{fig:syn2d}(a-i). Our goal is to construct a generative model that mimics these distributions. For an easy explanation, we rewrite the AC losses in Equations~\ref{eqn:cls_r} and \ref{eqn:cls_g} as
\begin{flalign}
  \label{eqn:cls_r_kl}
  {\cal L}_{\text{KL-AC}}^r & = \mathbb{E}_{({\bm x}^r, {\bm y}^r) \sim p^r({\bm x}, {\bm y})} {\cal D}_{\text{KL}} ({\bm y}^r \| {\bm s}^r),
  \\
  \label{eqn:cls_g_kl}
  {\cal L}_{\text{KL-AC}}^g & = \mathbb{E}_{{\bm z}^g \sim p^g({\bm z}), {\bm y}^g \sim p^g({\bm y})} {\cal D}_{\text{KL}} ({\bm y}^g \| {\bm s}^g),
\end{flalign}
where we rename the AC losses ${\cal L}_{\text{AC}}^r$ and ${\cal L}_{\text{AC}}^g$ as the KL-AC losses ${\cal L}_{\text{KL-AC}}^r$ and ${\cal L}_{\text{KL-AC}}^g$, respectively. Here, ${\cal D}_{\text{KL}}({\bm y} \| {\bm s}) = {\bm y}^{\top} \log {\bm y} - {\bm y}^{\top} \log {\bm s}$ is the Kullback-Leibler (KL) divergence between ${\bm y}$ and ${\bm s}$, where we represent the distributions in a vector form (\ie, ${\bm y}$ and ${\bm s}$). In practice, we ignore ${\bm y}^{\top} \log {\bm y}$ because it is constant when it is fixed (\eg, given as the ground truth).

Both the KL-AC losses bring ${\bm s}^r$ and ${\bm s}^g$ close to ${\bm y}^r$ and ${\bm y}^g$, respectively, in terms of the KL divergence; however, there is a difference for the optimization target. In ${\cal L}_{\text{KL-AC}}^r$, the input of $C$ (\ie, ${\bm x}^r$) is fixed and the parameters of $C$ are optimized. Namely, the class overlapping state in ${\bm x}^r$ is maintained during the training, as shown in Figure~\ref{fig:syn2d}(a-i). This encourages the classifier's posterior (${\bm s}^r$) to be optimized to represent the class overlapping state (Figure~\ref{fig:syn2d}(a-iii)) even when the ground truth class prior (${\bm y}^r$) is discrete (Figure~\ref{fig:syn2d}(a-ii)). In contrast, in ${\cal L}_{\text{KL-AC}}^g$, $C$ is fixed and the input of $C$ (\ie, ${\bm x}^g = G({\bm z}^g, {\bm y}^g)$) is optimized. Unlike in ${\cal L}_{\text{KL-AC}}^r$, there is no restriction that promotes $G$ to follow the class overlapping state. As a result, the classifier's posterior (${\bm s}^g$) is optimized to be close to the discrete prior (${\bm y}^g$; Figure~\ref{fig:syn2d}(b-ii)), as shown in Figure~\ref{fig:syn2d}(b-iii). This encourages the generative distribution (${\bm x}^g$) to be close to the separate state, as shown in Figure~\ref{fig:syn2d}(b-i). In this manner, AC-GAN prefers to learn separate class distributions even when classes overlap.\footnote{Refer to Shu~\etal~\cite{RShuNIPSW2017}, who explain these limitations of AC-GAN from a Lagrangian perspective.}

\subsection{Proposal: CP-GAN}
\label{subsec:cpgan}

To overcome these limitations, we redesign the generator input and the objective function of AC-GAN, as shown in Figure~\ref{fig:model}(b). First, to represent the between-class relationship in the generator input, we replace the discrete prior (${\bm y}^g$) with the classifier's posterior of real data (${{\bm s}}^r$). By this modification, in CP-GAN, the generator is formulated as ${\bm x}^g = G({\bm z}^g, {\bm s}^r)$.

Second, to render the classifier's posterior of the generated data (${\bm s}^g$) close to ${{\bm s}}^r$ instead of discrete ${\bm y}^g$, we reformulate the KL-AC loss of the generated data (Equation~\ref{eqn:cls_g_kl}) as
\begin{flalign}
  \label{eqn:cls_g_kl_cp}
  {\cal L}_{\text{KL-CP}}^g = \mathbb{E}_{{\bm z}^g \sim p^g({\bm z}), {\bm x}^r \sim p^r({\bm x})} {\cal D}_{\text{KL}} ({\bm s}^r \| {\bm s}^g),
\end{flalign}
where ${\bm s}^r = C({\bm y} | {\bm x}^r)$ and ${\bm s}^g = C({\bm y} | G({\bm z}^g, {\bm s}^r))$. We rename the KL-AC loss as the KL-CP loss. By minimizing this loss, $G$ is encouraged to generate data of which classifier's posterior (${\bm s}^g$) is close to that of real data (${\bm s}^r$) in terms of the KL divergence.

An advantage of CP-GAN is that the distribution shape of its generator prior (\ie, ${\bm s}^r$) is determined in a data-driven manner. When real data has class overlaps (Figure~\ref{fig:syn2d}(a-i)), ${\bm s}^r$ represents the class overlapping state (Figure~\ref{fig:syn2d}(a-iii)). This allows $G$ to generate class-overlapping data (Figure~\ref{fig:syn2d}(c-i)). When real data is discrete, ${\bm s}^r$ also becomes discrete, \ie, close to ${\bm y}^g$. In this case, CP-GAN is close to AC-GAN.

Another advantage is that CP-GAN does not require an additional network, compared with AC-GAN (Figures~\ref{fig:model}(a) and (b)). Additionally, CP-GAN does not require extra supervision. This means that the calculation cost and the annotation cost are almost the same between AC-GAN and CP-GAN.

\smallskip\noindent
\textbf{Prior GAN (pGAN).}
As shown in Figure~\ref{fig:model}(b), when generating an image, CP-GAN requires ${\bm x}^r$ to calculate the generator prior ${\bm s}^r = C({\bm y} | {\bm x}^r)$. To alleviate this requirement, we additionally introduce \textit{prior GAN (pGAN)} that mimics the distribution of ${\bm s}^r$. As shown in Figure~\ref{fig:model}(c), pGAN has the cGAN formulation~\cite{MMirzaArXiv2014} and the objective function is defined as
\begin{eqnarray}
  \label{eqn:pGAN}
  {\cal L}_{\text{pGAN}} \! = \! \mathbb{E}_{({\bm x}^r, {\bm y}^r) \sim p^r({\bm x}, {\bm y})} [ \log D_p({\bm s}^r, {\bm y}^r) ] \! + \! \mathbb{E}_{{\bm z}^g_p \sim p^g_p({\bm z}), {\bm y}^g_p \sim p^g_p({\bm y})} [ \log (1 \! - \! D_p(G_p({\bm z}^g_p, {\bm y}^g_p), {\bm y}^g_p)) ],
\end{eqnarray}
where $G_p$ and $D_p$ are a generator and a discriminator, respectively; ${\bm z}^g_p$ is the noise sampled from a normal distribution $p^g_p({\bm z}) = {\cal N}(0, I)$ and ${\bm y}^g_p$ is the class label sampled from a categorical distribution $p^g_p({\bm y}) = \text{Cat}(K=c, p=1/c)$. When generating an image using pGAN and CP-GAN, we first generates a classifier posterior ${\bm s}^g_p = G_p({\bm z}^g_p, {\bm y}^g_p)$ using pGAN and subsequently generates an image ${\bm x}^g = G({\bm z}^g, {\bm s}^g_p)$ using CP-GAN. Note that ${\bm s}^r$ is simpler than ${\bm x}^r$; therefore, it is relatively easy to learn a generator that mimics ${\bm s}^r$. In Section~\ref{sec:experiments}, we empirically find that almost no degradation occurs by using $G_p({\bm z}^g_p, {\bm y}^g_p)$ instead of ${\bm s}^r$.

\subsection{Relationships with previous class-conditional extensions of GANs}
\label{subsec:relationship}

In Table~\ref{tab:relationships}, we summarize the relationships between CP-GAN and previous class-conditional extensions of GANs, including typical models (\ie, cGAN with a concat discriminator~\cite{MMirzaArXiv2014} and AC-GAN~\cite{AOdenaICML2017}) and state-of-the-art extensions (\ie, cGAN with a projection discriminator~\cite{TMiyatoICLR2018} and conditional filtered GAN (CFGAN)~\cite{TKanekoCVPR2017}). Our technical contributions are three-fold. (1) We incorporate the classifier's posterior into the generator input to represent the between-class relationships. (2) We redesign the discriminator objective function and introduce the KL-CP loss to make the classifier's posterior of generated data correspond with that of real data. (3) We additionally introduce pGAN that mimics the classifier's posterior of real data. As demonstrated in Section~\ref{sec:experiments}, these modifications allow the generator to capture the between-class relationships and generate an image conditioned on the class specificity even where the previous class-conditional extensions of GANs fail to do so.

\begin{table*}[htb]
  \centering
  \scalebox{0.75}[0.75]{
    \begin{tabular}{c|ll|ll}
      \bhline{1pt}
      \multirow{2}{*}{{\bf Model}}
      & \multicolumn{2}{c|}{{\bf Generator}}
      & \multicolumn{2}{c}{{\bf Discriminator}} \\ \cline{2-5}
      & {\bf Formulation} & {\bf Input}
      & {\bf Formulation} & {\bf Objective} \\ \hline
      AC-GAN~\cite{AOdenaICML2017}
      & $G({\bm z}^g, {\bm y}^g)$ & ${\bm y}^g \sim \text{Cat}$
      & $D({\bm x})$ with $C({\bm y}|{\bm x})$ & AC loss \\
      cGAN (concat)~\cite{MMirzaArXiv2014}
      & $G({\bm z}^g, {\bm y}^g)$ & ${\bm y}^g \sim \text{Cat}$
      & $D({\bm x}, {\bm y})$ & Concat $D$ \\
      cGAN (projection)~\cite{TMiyatoICLR2018}
      & $G({\bm z}^g, {\bm y}^g)$ & ${\bm y}^g \sim \text{Cat}$
      & $D({\bm x}, {\bm y})$ & Projection $D$ \\      
      CFGAN~\cite{TKanekoCVPR2017}\dag
      & $G({\bm z}^g, {\bm m}^g)$
      & ${\bm m}^g \sim \text{Mix}$ cond. on ${\bm y}^g \sim \text{Cat}$
      & $D({\bm x}, {\bm y})$ with $Q({\bm m}|{\bm x}, {\bm y})$
                          & Cond. MI loss \\ \hline
      CP-GAN
      & $G({\bm z}^g, {\bm s})$
      & ${\bm s} = C({\bm y} | {\bm x}^r)$
        (or ${\bm s} = G_p({\bm z}^g_p, {\bm y}^g_p)$)\!
      & $D({\bm x})$ with $C({\bm y}|{\bm x})$ & KL-CP loss \\
      \bhline{1pt}          
    \end{tabular}        
  }
  \vspace{1mm}
  \caption{Relationships between CP-GAN and previous class-conditional extensions of GANs. \dag Naive CFGAN assumes that the number of classes is only two. For comparison purposes, we extend this to class mixture setting.}
  \label{tab:relationships}
  \vspace{-3mm}
\end{table*}

\section{Experiments}
\label{sec:experiments}

In this section, we empirically verify the proposed model on controlled class-overlapping data in Section~\ref{subsec:exp_controlled} and real class-overlapping data in Section~\ref{subsec:exp_real}. Due to the space limitation, we briefly review the experimental setup and only provide the important results in this main text. See the Appendix\footnote{In Appendix~\ref{sec:exp_pix2pix}, we demonstrate the generality of the proposed model by applying it to image-to-image translation. Appendices~\ref{sec:extended} and \ref{sec:details} provides more results and details of experimental setup, respectively.} and our \href{https://takuhirok.github.io/CP-GAN/}{website} for details and more results.

\vspace{-2mm}
\subsection{Experiments on controlled class-overlapping data}
\label{subsec:exp_controlled}

\vspace{-1mm}
\subsubsection{Experimental setup}
\label{subsubsec:exp_setup}
\begin{figure*}[t]
  \centering
  \includegraphics[width=0.99\textwidth]{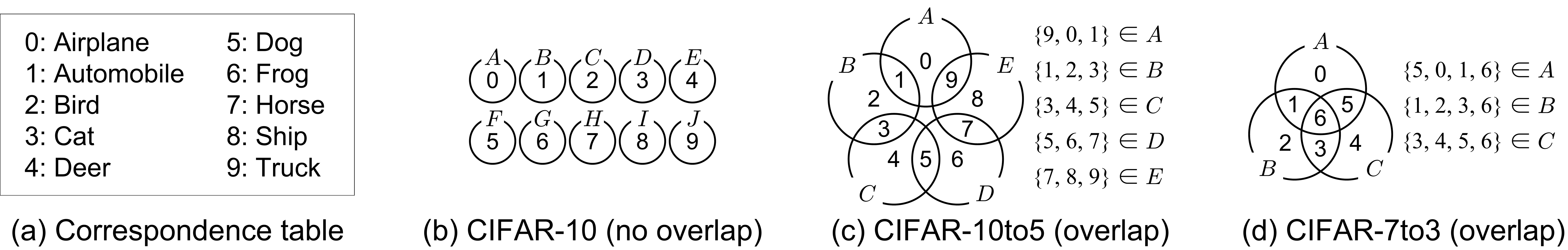}
  \vspace{-2mm}
  \caption{Illustration of class overlapping settings. (a) $0, 1, \dots$ represent \textit{airplane}, \textit{automobile}, $\dots$, respectively. (b) This is the original CIFAR-10 dataset (no overlap). In (c) and (d), we divided the original \textit{ten} and \textit{seven} classes into \textit{five} and \textit{three} classes, respectively, following class labels denoted by alphabet letters ($A, B, \dots$). During the training, we can only use these alphabet class labels and cannot observe the digit class labels ($0, 1, \dots$).}
  \label{fig:overlap}
\end{figure*}

\textbf{Dataset.}
To examine the benchmark performance on our novel task (class-distinct and class-mutual image generation), we used CIFAR-10~\cite{AKrizhevskyTech2009} in which we constructed the class overlapping states in a controlled manner.\footnote{We chose this dataset because it is commonly used both in image generation~\cite{MLucicNeurIPS2018,IGulrajaniNIPS2017,TMiyatoICLR2018b} and in noisy label image classification~\cite{CZhangICLR2017,DArpitICML2017}. For the latter task, class labels are corrupted in a controlled manner, as here.} We consider three settings to cover various situations. (1) \textbf{CIFAR-10} (Figure~\ref{fig:overlap}(b)): This is the original CIFAR-10 dataset. We used this setting to confirm whether CP-GAN does not cause a negative effect even in class non-overlapping data. (2) \textbf{CIFAR-10to5} (Figure~\ref{fig:overlap}(c)): We divided the original \textbf{ten} classes ($0, \dots, 9$; defined in Figure~\ref{fig:overlap}(a)) into \textbf{five} classes ($A, \dots, E$) with class overlapping. In this setting, class overlapping exists between partial two-class pairs. Hence, classes may overlap (\eg, between $A$ and $B$) or not overlap (\eg, between $A$ and $C$). (3) \textbf{CIFAR-7to3} (Figure~\ref{fig:overlap}(d)): We divided the original \textbf{seven} classes ($0, \dots, 6$) into \textbf{three} classes ($A, B, C$) with class overlapping. In this setting, we need to handle three different states: a class-distinct state (\eg, $A \rightarrow 0$), two-class overlap state (\eg, $A \cap B \rightarrow 1$), and three-class overlap state (\eg, $A \cap B \cap C \rightarrow 6$).

\smallskip\noindent
\textbf{Evaluation metrics.}
We used two evaluation metrics for comprehensive analysis, \ie, class-distinct and class-mutual accuracy (\textbf{DMA}) and Fr\'{e}chet inception distance (\textbf{FID})~\cite{MHeuselNIPS2017}. (1) \textbf{DMA:} To confirm whether a model can generate class-distinct and class-mutual images selectively, we introduced DMA. We first generate an image conditioned on either a class-distinct or class-mutual state, and calculate the accuracy for the expected state. For example, in \textbf{CIFAR-10to5} (Figure~\ref{fig:overlap}(c)), the class-distinct state for $A$ is represented by ${\bm y}_A$, where ${\bm y}_X$ is a one-hot vector denoting the class $X$. When an image is generated from ${\bm y}_A$, the image is expected to be classified as class $0$ (\textit{airplane}). In contrast, the class-mutual state for $A \cap B$ is represented by $({\bm y}_A + {\bm y}_B) / 2$. When an image is generated from $({\bm y}_A + {\bm y}_B) / 2$, the image is expected to be classified as class $1$ (\textit{automobile}). We compute this accuracy for all class-distinct and class-mutual states and report the average accuracy over all states. (2) \textbf{FID:} To assess the generated image quality, we used the FID that measures the distance between $p^r$ and $p^g$ in Inception embeddings.\footnote{Another typical metric is the Inception score (IS)~\cite{TSalimansNIPS2016}. However, its drawbacks have been indicated by the recent studies~\cite{MHeuselNIPS2017,MLucicNeurIPS2018}. Therefore, we used the FID in this study.}

\subsubsection{Analysis of model configurations}
\label{subsubsec:exp_config}

\textbf{Generalization and memorization.}
A DNN classifier can memorize even noisy (or random) labels~\cite{CZhangICLR2017}. This is critical for CP-GAN because when $C$ memorizes labels completely, it is difficult to capture between-class relationships. However, appropriately tuned explicit regularization (\eg, dropout) can degrade the training performance on noisy data without compromising generalization on real data~\cite{DArpitICML2017}. This indicates that it may be possible to learn a well-regularized $C$ in CP-GAN by configuring a model appropriately. To detail this problem, we analyzed model configurations while focusing on the components that affect the regularization of $C$. We examined three aspects: (1) The effect of \textbf{explicit regularization} (\ie, dropout). (2) The effect of a \textbf{shared architecture} between $D$ and $C$. (3) The effect of \textbf{joint training} between ${\cal L}_{\text{GAN}}$ and ${\cal L}_{\text{KL-CP}}^g$. We implemented the models based on WGAN-GP ResNet~\cite{IGulrajaniNIPS2017} as commonly used benchmarks.

\begin{table}[tb]  
  \centering
  \scalebox{0.6}[0.6]{
    \begin{tabular}{c|c|c|cc|cc|cc}
      \bhline{1pt}
      \multirow{2}{*}{\textbf{No.}}
      & \multicolumn{2}{c|}{\multirow{2}{*}{\textbf{Condition}}}
      & \multicolumn{2}{c|}{\textbf{CIFAR-10}}
      & \multicolumn{2}{c|}{\textbf{CIFAR-10to5}}
      & \multicolumn{2}{c}{\textbf{CIFAR-7to3}}
      \\ \cline{4-9}
      & \multicolumn{2}{c|}{}
      & \!{\textbf{DMA}} $\uparrow$\! & \!{\textbf{FID}} $\downarrow$\!
      & \!{\textbf{DMA}} $\uparrow$\! & \!{\textbf{FID}} $\downarrow$\!
      & \!{\textbf{DMA}} $\uparrow$\! & \!{\textbf{FID}} $\downarrow$\!
      \\ \hline
      
      1 & \multirow{2}{*}{Dropout}
      & W/ & 95.4 & 11.6 & 77.4 & 12.2 & 60.1 & 13.6
      \\
      2 & & W/o & 94.8 & 12.7 & 51.7 & 14.9 & 37.9 & 18.4      
      \\ \hline
      3 & \multirow{4}{*}{\# of shared blocks}
      & 0 & 96.8 & 20.3 & 56.7 & 20.9 & 43.3 & 23.9
      \\
      4 & & 1 & 96.5 & 19.0 & 61.0 & 19.9 & 40.1 & 22.4
      \\
      5 & & 2 & 94.8 & 15.3 & 72.6 & 15.2 & 33.9 & 35.7
      \\
      6 & & 3 & 95.7 & 13.0 & 78.1 & 13.6 & 59.0 & 16.7
      \\ \hline      
      7 & \multirow{2}{*}{Iter. of adding ${\cal L}_{\text{KL-CP}}^g$}
      & $20k$ & 95.0 & 12.3 & 71.2 & 13.5 & 49.9 & 15.4
      \\
      8 & & $40k$ & 95.8 & 12.5 & 63.8 & 15.0 & 46.8 & 14.8
      \\ \hline
      9 & \multirow{3}{*}{$\lambda^g$}
      & 0.2 & 96.8 & 11.2 & 88.7 & 12.4 & 72.2 & 14.5
      \\
      10 & & 0.4 & 97.1 & 12.0 & 95.0 & 12.5 & 82.0 & 14.6
      \\
      11 & & 1 & 97.4 & 13.8 & 96.8 & 13.9 & 91.1 & 15.3
      \\ \bhline{1pt}
    \end{tabular}
  }
  \vspace{1mm}
  \caption{Analysis of model configurations. No.~1 is the default model (with dropout, four shared blocks, using ${\cal L}_{\text{KL-CP}}^g$ from the beginning, and $\lambda_g = 0.1$). In the other models, only the target parameters are changed. In DMA, the larger the value, the better. In FID, the smaller the value, the better.}
  \label{tab:eval_cifar10}
\end{table}

Nos.~1--8 of Table~\ref{tab:eval_cifar10} show the results. (1) In the comparison with and without dropout (nos.~1--2), dropout helps to improve both DMA and FID. Regularization by dropout may prevent the models from converging to a few modes (\eg, only class-distinct states) and allow $G$ to capture both class-distinct and class-mutual states. (2) When comparing the number of shared blocks (nos.~1, 3--6), DMA and FID tend to become better (or comparable) as the shared number increases. This indicates that the shared architecture also acts as an effective regularizer. (3) When comparing timing of using ${\cal L}_{\text{KL-CP}}^g$ (nos.~1, 7--8), joint training in an early stage (in which $C$ prioritizes learning a simple pattern \cite{DArpitICML2017}) helps to improve DMA. Through this analysis, we confirm that explicit regularization, shared architecture, and joint training from the beginning are useful for regularizing $C$ and preventing the memorization.

\smallskip\noindent
\textbf{Effect of trade-off parameter ${\lambda}^g$.}
Another important parameter that affects the performance is ${\lambda}^g$, which weights the importance between the adversarial loss and the KL-CP loss. Nos.~1, 9--11 of Table~\ref{tab:eval_cifar10} compare the results for ${\lambda}^g$. We observe a trade-off: the larger ${\lambda}^g$ improves the DMA but degrades the FID. However, we find that in class-overlapping cases (\ie, CIFAR-10to5 and CIFAR-7to3), the degradation of the FID is relatively small in ${\lambda}^g \in [0.1, 0.4]$ while improving the DMA by a large margin.

\smallskip\noindent
\textbf{Effect of pGAN.}
As discussed in Section~\ref{subsec:cpgan}, CP-GAN requires a real ${\bm x}^r$ to obtain the generator input ${\bm s}^r$. To mitigate this requirement, we introduced pGAN that instead generates ${\bm s}^r$ using randomly sampled variables (\ie, ${\bm z}^g_p$ and ${\bm y}^g_p$). Here, we examined the performance of the combination of pGAN and CP-GAN. In particular, we used CP-GAN listed in no.~1 of Table~\ref{tab:eval_cifar10}. The FID scores for CIFAR-10, CIFAR-10to5, and CIFAR-7to3 are 11.7, 12.2, and 13.7, respectively. They are comparable to the scores in no.~1 of Table~\ref{tab:eval_cifar10} (the difference is within 0.1). See Figure~\ref{fig:cifar10_cm} in the Appendix, which shows the comparison of ${\bm s}^r$ (real classifier's posterior) and ${\bm s}^g_p$ (generated classifier's posterior using pGAN). This figure also confirms that pGAN can mimic a real classifier's posterior reasonably well.

\subsubsection{Comparison with previous class-conditional extensions of GANs}
\label{subsubsec:exp_comparison}

To verify the effectiveness of CP-GAN, we compared it with three models discussed in Section~\ref{subsec:relationship}: (1) \textbf{AC-GAN}~\cite{AOdenaICML2017}; (2) \textbf{cGAN} with a projection discriminator \cite{TMiyatoICLR2018}, which uses $D({\bm x}, {\bm y})$ instead of the combination of $D({\bm x})$ and $C({\bm y} | {\bm x})$; (3) \textbf{CFGAN}~\cite{TKanekoCVPR2017}, with a manually defined class-mixture prior as the generator prior. Unlike CP-GAN, CFGAN requires defining the class-mixture method in advance. We mixed two classes and three classes uniformly for CIFAR-10to5 and CIFAR-7to3, respectively. We did not test CFGAN on CIFAR-10 because this dataset has no class overlaps. We implemented the models based on \textbf{WGAN-GP} ResNet~\cite{IGulrajaniNIPS2017}.\footnote{Based on the findings in Table~\ref{tab:eval_cifar10}, we apply dropout to all models and fix ${\lambda}^g = 0.4$ in AC-GAN and CP-GAN.} For fair comparison, we also tested \textbf{SN-GAN} ResNet~\cite{TMiyatoICLR2018b}, which is the optimal GAN configuration for \textbf{cGAN}.

\begin{figure*}[tb]
  \def\@captype{table}
  \begin{minipage}[t]{.57\textwidth}
    \centering
    \scalebox{0.6}[0.6]{
      \begin{tabular}{c|c|cc|cc|cc}
        \bhline{1pt}
        \multirow{2}{*}{\!\textbf{Configuration}\!}
        & \multirow{2}{*}{\textbf{GAN}}
        & \multicolumn{2}{c|}{\textbf{CIFAR-10}}
        & \multicolumn{2}{c|}{\textbf{CIFAR-10to5}}
        & \multicolumn{2}{c}{\textbf{CIFAR-7to3}}
        \\ \cline{3-8}
        &
        & \!{\textbf{DMA}} $\uparrow$\! & \!{\textbf{FID}} $\downarrow$\!
        & \!{\textbf{DMA}} $\uparrow$\! & \!{\textbf{FID}} $\downarrow$\!
        & \!{\textbf{DMA}} $\uparrow$\! & \!{\textbf{FID}} $\downarrow$\!
        \\ \hline
        \multirow{4}{*}{WGAN-GP}
        & \!AC-GAN\! & 94.9 & 12.5 & 36.6 & 13.7 & 30.2 & 14.7
        \\
        & \!cGAN\! & 81.0 & 15.9 & 32.3 & 16.9 & 26.2 & 18.8
        \\
        & \!CFGAN\! & -- & -- & 50.9 & 15.8 & 43.0 & 16.8
        \\
        & \!CP-GAN\! & \textbf{97.1} & \textbf{12.0} & \textbf{95.0} & \textbf{12.5}
        & \textbf{82.0} & \textbf{14.6}
        \\ \hline
        \multirow{4}{*}{SN-GAN}
        & \!AC-GAN\! & 90.4 & 12.7 & 31.4 & 13.6 & 30.5 & 15.5
        \\
        & \!cGAN\! & 87.6 & \textbf{10.8} & 36.2 & 13.7 & 27.5 & 16.6
        \\
        & \!CFGAN\! & -- & -- & 54.5 & 16.4 & 43.1 & 20.8
        \\
        & \!CP-GAN\! & \textbf{95.2} & 11.4 & \textbf{64.9} & \textbf{13.0} & \textbf{51.6} & \textbf{15.2}
        \\ \bhline{1pt}
      \end{tabular}
    }    
    \vspace{1mm}
    \tblcaption{Comparing class-conditional extensions of GANs in controlled class-overlapping data.}
    \label{tab:compare_controlled}
  \end{minipage}
  \hfill  
  \begin{minipage}[c]{.4\textwidth}
    \centering
    \includegraphics[width=0.98\textwidth]{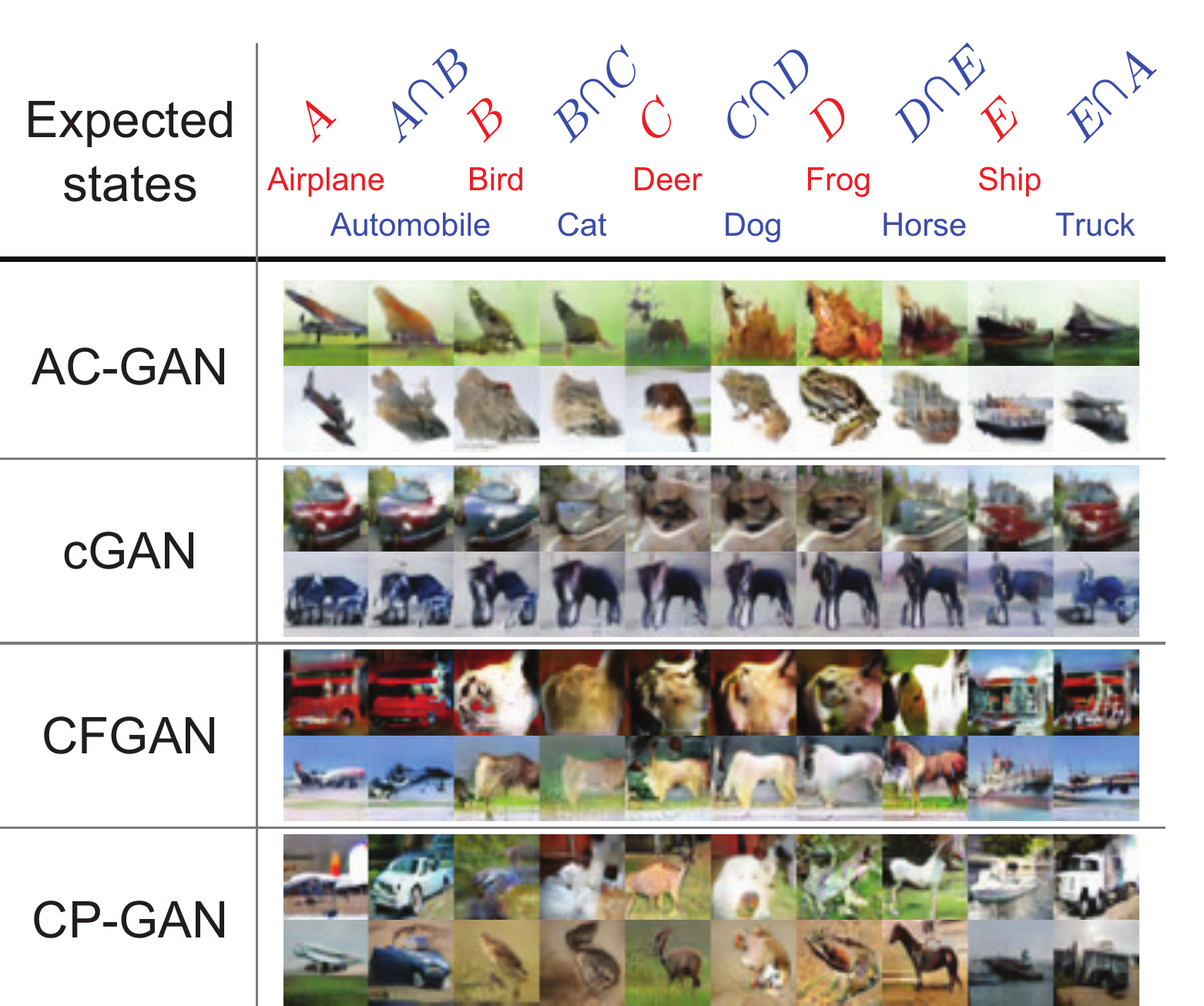}
    \vspace{-2mm}
    \caption{Samples of generated images on CIFAR-10to5. See Figures~\ref{fig:cifar10_gen_ex} and \ref{fig:cifar10_interp} in the Appendix for more results.}
    \label{fig:cifar10_gen}
  \end{minipage}
  \vspace{-2mm}
\end{figure*}

\smallskip\noindent
\textbf{Results.}
We list the results in Table~\ref{tab:compare_controlled}. When using WGAN-GP, CP-GAN outperforms the other models in terms of both DMA and FID in all datasets. When using SN-GAN, CP-GAN outperforms the other models in terms of the DMA in all datasets. As of the FID, cGAN achieves the best performance in CIFAR-10. This coincides with the results in \cite{TMiyatoICLR2018}. However, we find that it does not necessarily work best in class-overlapping data. Thus, it is important to consider class-distinct and class-mutual image generation separately from typical discrete class-conditional image generation. Figure \ref{fig:cifar10_gen} shows the generated images. CP-GAN succeeds in selectively generating class-distinct (red font) and class-mutual (blue font) images corresponding with the expected states, whereas the other models fail to do so.

\subsection{Experiments on real-world class-overlapping data}
\label{subsec:exp_real}

We tested CP-GAN on Clothing1M~\cite{TXiaoCVPR2015}, which consists of 14-class clothing images. The data are collected from shopping websites in a real-world scenario and include many mislabels resulting from confusion, as shown in Figure~\ref{fig:example}(d) (the annotation accuracy is $61.54\%$). This dataset contains $1M$ \textit{noisy} and $50k$ \textit{clean} labeled data as training sets. We used only \textit{noisy} labeled data to address the most realistic setting. Our goal is to construct a generative model that can capture class mixture resulting from confusion and generate class-specific and class-ambiguous images selectively. Based on the findings in Section~\ref{subsec:exp_controlled}, we implemented CP-GAN based on WGAN-GP ResNet~\cite{IGulrajaniNIPS2017} and applied dropout (we call this configuration WGAN-GP ResNet-Dropout). We also tested two models as baselines: (1) \textbf{AC-GAN} \cite{AOdenaICML2017} (with WGAN-GP ResNet-Dropout) (2) \textbf{cGAN} with a projection discriminator \cite{TMiyatoICLR2018} (with SN-GAN ResNet).\footnote{We also tested cGAN with WGAN-GP ResNet-Dropout, but we found that it suffers from severe mode collapse during training. Hence, we instead used SN-GAN ResNet, which is an optimal configuration for cGAN.} To shorten the training time, we resized images from $256 \times 256$ to $64 \times 64$. In this dataset, class-mutual states are not given as ground truth; therefore, it is difficult to calculate the DMA. Hence, we instead reported class-distinct accuracy (\textbf{DA}), \ie, we calculated the accuracy only for the class-distinct states (\ie, ${\bm y}^g$ is a one-hot vector).

\begin{figure*}[tb]
  \def\@captype{table}
  \begin{minipage}[t]{.3\textwidth}
    \centering
    \scalebox{0.6}[0.6]{
      \begin{tabular}{c|cc}
        \bhline{1pt}
        {\textbf{GAN}}
        & {\textbf{DA}} $\uparrow$ & {\textbf{FID}} $\downarrow$
        \\ \hline
        AC-GAN & 46.3 & 9.3
        \\
        cGAN & 49.5 & 11.4
        \\
        CP-GAN & \textbf{65.1} & \textbf{6.8}
        \\ \bhline{1pt}
      \end{tabular}
    }
    \vspace{1mm}
    \tblcaption{Comparing class-conditional extensions of GANs in real-world class-overlapping data.}
    \label{tab:compare_real}
  \end{minipage}
  \hfill
  \begin{minipage}[c]{.67\textwidth}
    \centering
    \includegraphics[width=0.98\textwidth]{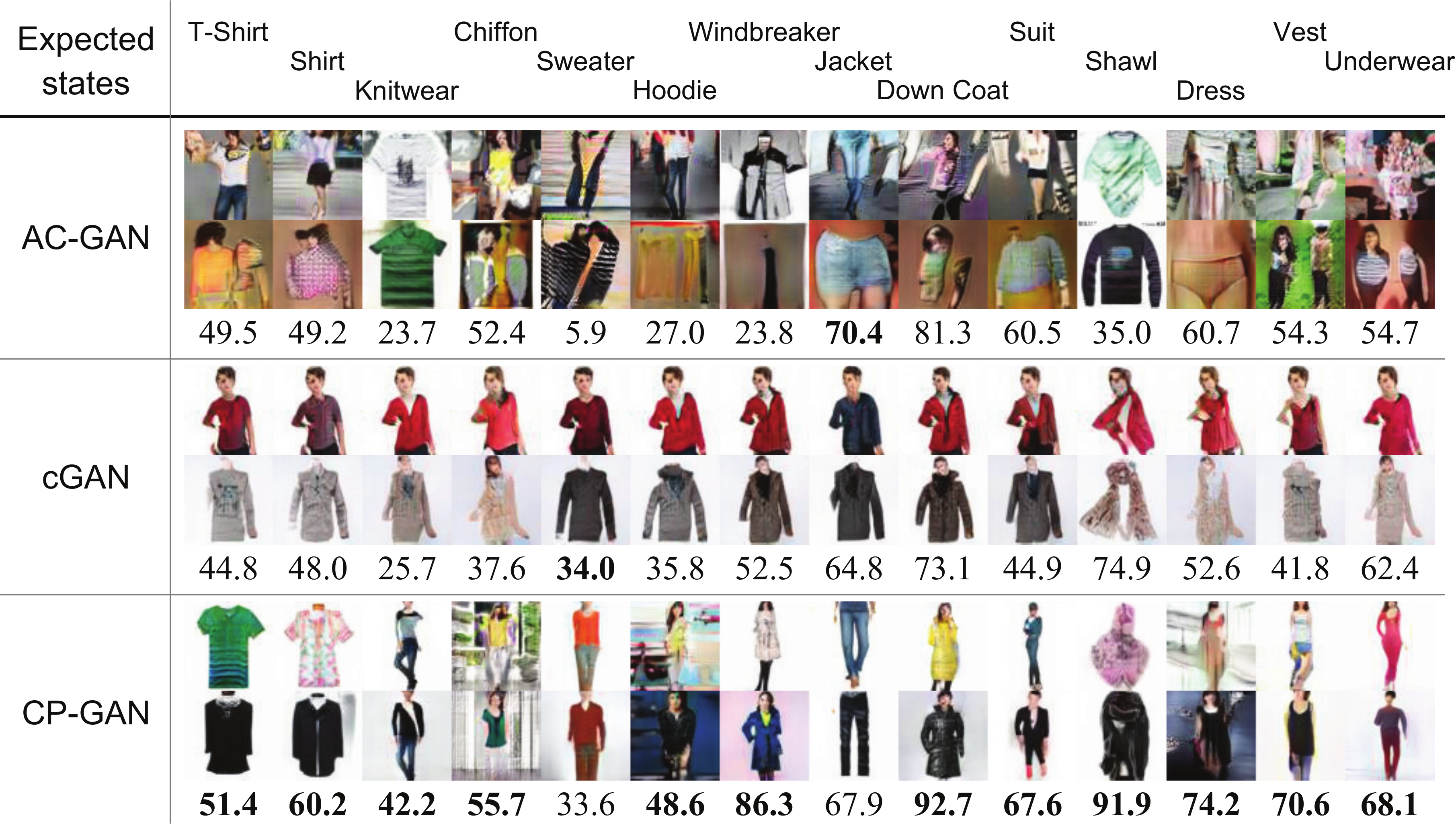}
    \vspace{-2mm}
    \caption{Samples of generated images on Clothing1M. Number below images indicates the per-class DA. See Figure~\ref{fig:clothing1m_gen_ex} in the Appendix for more results.}
    \label{fig:clothing1m_gen}
  \end{minipage}
  \vspace{-2mm}
\end{figure*}

\smallskip\noindent
\textbf{Results.}
We list the results in Table~\ref{tab:compare_real}. We confirm that CP-GAN works better than AC-GAN and cGAN in terms of DA and FID even in real-world class-overlapping data. These results indicate that class-overlapping data could cause learning difficulties in previous class-conditional models, and it is important to incorporate a mechanism like ours. We present the qualitative results and per-class DA in Figure~\ref{fig:clothing1m_gen}. CP-GAN succeeds in generating the most class-distinct images in terms of the per-class DA in most cases (12/14). We tested the combination of CP-GAN and pGAN also in this dataset. This achieves the FID of 6.8, which is the same as the score listed in Table~\ref{tab:compare_real}. See Figure~\ref{fig:clothing1m_cm} in the Appendix, which shows that pGAN (${\bm s}^g_p$) can mimic ${\bm s}^r$ reasonably well.

\section{Conclusions}
\label{sec:conclusions}

This study introduced a new problem called class-distinct and class-mutual image generation, in which we aim to construct a generative model that can be controlled for class specificity. To solve this problem, we redesigned the generator input and the objective function of AC-GAN, and developed CP-GAN that solves this problem without additional supervision. In the experiments, we demonstrated the effectiveness of CP-GAN using both controlled and real-world class-overlapping data along with a model configuration analysis and a comparative study with state-of-the-art class-conditional extensions of GANs. Based on our findings, adapting our method to other generative models such as VAEs~\cite{DKingmaICLR2014,DRezendeICML2014}, ARs~\cite{AOordICML2016}, and Flows~\cite{LDinhICLR2017} and using it as a data-mining tool on real-world complex datasets remain interesting future directions.

\smallskip\noindent
\textbf{Acknowledgement.}
We would like to thank Hiroharu Kato, Atsuhiro Noguchi, and Antonio Tejero-de-Pablos for helpful discussions. This work was supported by JSPS KAKENHI Grant Number JP17H06100, partially supported by JST CREST Grant Number JPMJCR1403, Japan, and partially supported by the Ministry of Education, Culture, Sports, Science and Technology (MEXT) as ``Seminal Issue on Post-K Computer.''

\bibliography{egbib}

\clearpage
\appendix
\section{Application to image-to-image translation}
\label{sec:exp_pix2pix}

Our proposed models are general extensions of AC-GAN. Therefore, they can be incorporated into any AC-GAN-based model. To demonstrate this, we incorporate CP-GAN into StarGAN~\cite{YChoiCVPR2018}, which is a model for multi-domain image-to-image translation. StarGAN is optimized using the AC loss with the adversarial loss~\cite{IGoodfellowNIPS2014} and cycle-consistency loss~\cite{JYZhuICCV2017}. To combine CP-GAN with StarGAN, we alter the class representation from discrete (${\bm y}^g$) to classifier's posterior (${\bm s}^r$), and replace the AC loss with the KL-CP loss. We call the combined model \textbf{CP-StarGAN}. To validate the effectiveness, we employ a modified version of CelebA~\cite{ZLiuICCV2015} in which we consider the situation where multiple datasets collected on diverse criteria are given. In particular, we divided CelebA into three subsets: ($A$) a \textit{black hair} set, ($B$) \textit{male} set, and ($C$) \textit{smiling} set. Our goal is to discover class-distinct (\eg, $A \cap \neg B$: \textit{back hair} and \textit{not male}) and class-mutual (\eg, $A \cap B$: \textit{back hair} and \textit{male}) representations without relying on any additional supervision. We believe that this would be useful for a real-world application in which we want to discover class intersections for existing diverse datasets.

\begin{figure*}[ht]
  \centering
  \vspace{-1mm}
  \includegraphics[width=\textwidth]{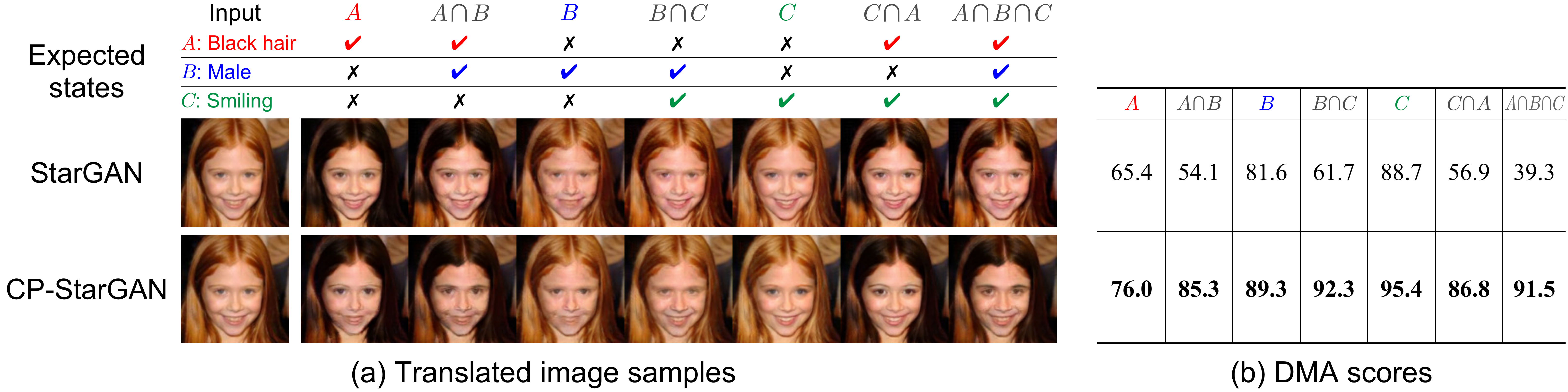}
  \vspace{-5mm}
  \caption{Samples of translated image and DMA scores on CelebA. We calculate the DMA scores for each class-distinct or class-mutual state. During the training, we only know the dataset identifiers indicated by colorful font, \ie, $A$ (\textit{black hair}), $B$ (\textit{male}), and $C$ (\textit{smiling}). Class-distinct and class-mutual representations (\eg, $A \cap B$; \textit{black hair} and \textit{male}) are found through learning. See Figure~\ref{fig:pix2pix_gen_ex} for more samples.}
  \vspace{-1mm}
  \label{fig:pix2pix_gen}
\end{figure*}

\smallskip\noindent\textbf{Results.}
We present the translated image samples and quantitative evaluation results
in Figure~\ref{fig:pix2pix_gen}.\footnote{To calculate DMA, we trained three classifiers (which distinguish male or not, black hair or not, and smiling or not, respectively). We calculated the accuracy for the expected states using these classifiers and reported their average score.} These results imply that CP-StarGAN can selectively generate class-distinct (\eg, $A \cap \neg B \cap \neg C$: \textit{black hair}, \textit{not male}, and \textit{not smiling}) and class-mutual (\eg, $A \cap B \cap C$: \textit{black hair}, \textit{male}, and \textit{smiling}) images accurately, whereas conventional StarGAN fails. Note that such multi-dimensional attribute representations (\eg, \textit{black hair}, \textit{not male}, and \textit{not smiling} ($A \cap \neg B \cap \neg C$)) are not given as supervision during the training. Instead, only the dataset identifiers (\ie, $A$, $B$, or $C$) are given as supervision, and we discover the multi-dimensional representations through learning.

\smallskip\noindent
\textbf{Related work.}
Recently, class-conditional extensions of GANs were applied to not only to image generation but also to image-to-image translation~\cite{YChoiCVPR2018,ZHeArXiv2017,BZhaoECCV2018,ARomeroArXiv2018} (including StarGAN~\cite{YChoiCVPR2018}). Their goal is to achieve multi-domain image-to-image translation (\ie, to obtain mappings among multiple domains) using few-parameter models. To achieve this, they use the AC-GAN-based loss. Therefore, they suffer from difficulty in applying to class-overlapping data like typical class-conditional image generation models. However, this difficulty can be overcome by replacing the AC-GAN-based loss with CP-GAN-based loss, as shown above.

\clearpage
\section{Extended and additional results}
\label{sec:extended}

\subsection{Extended results of Section~\ref{subsec:exp_controlled}}
\label{subsec:ex_cifar10}

\begin{figure*}[ht]
  \centering  
  \includegraphics[width=1\linewidth]{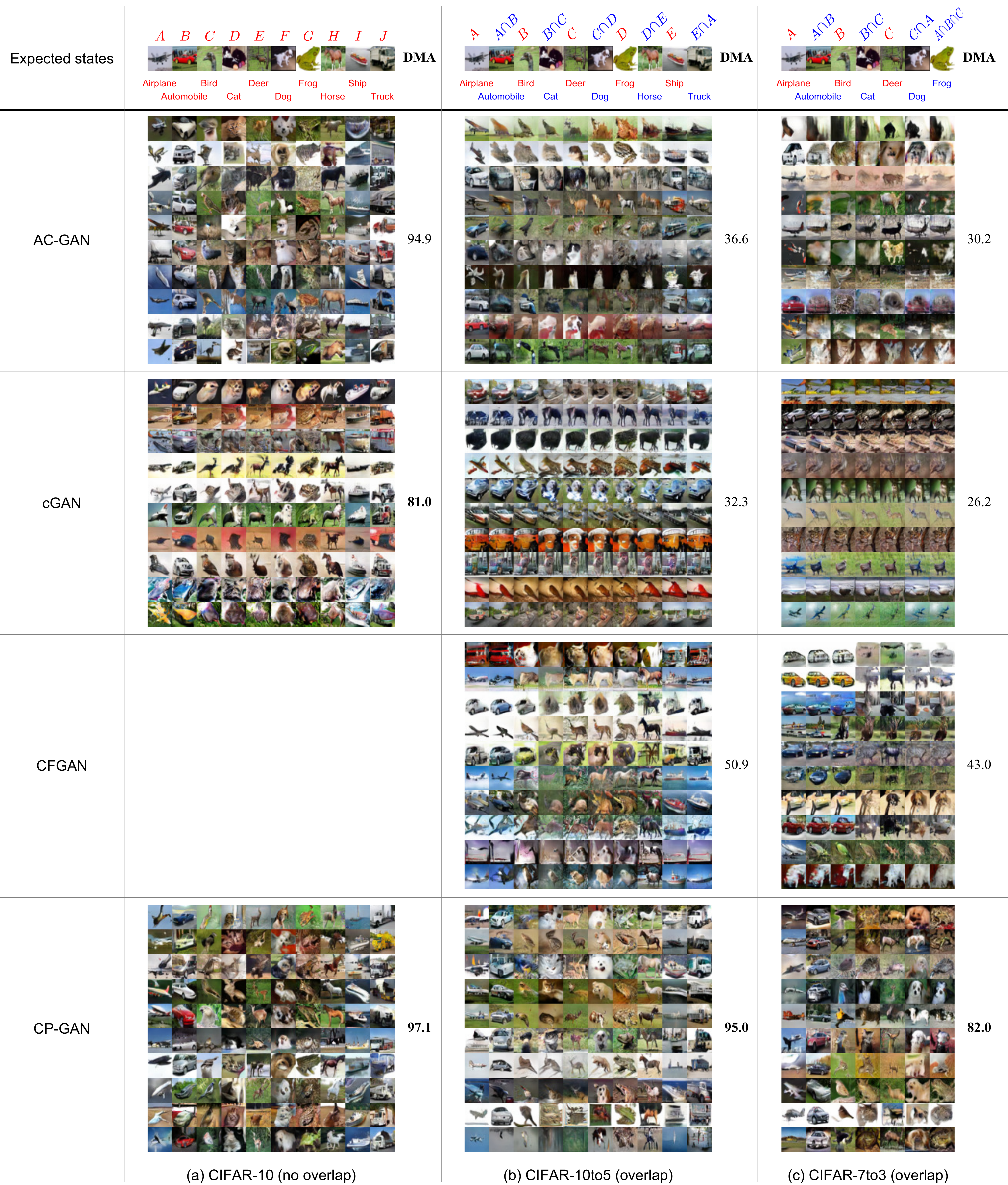}
  \caption{Samples of generated images on CIFAR-10, CIFAR-10to5, and CIFAR-7to3. This is the extended version of Figure~\ref{fig:cifar10_gen}. Each row shows samples generated with a fixed ${\bm z}^g$ and a varied ${\bm y}^g$. Each column contains samples generated with the same ${\bm y}^g$. CP-GAN succeeds in selectively generating class-distinct (red font) and class-mutual (blue font) images that correspond with the expected states (which are shown in the first row), whereas AC-GAN, cGAN, and CFGAN fail to do so. Note that we do not use class labels of categories (\textit{airplane}, \textit{automobile}, $\dots$) directly as supervision. Instead, we derive them from class labels denoted by alphabet letters ($A, B, \dots$). Refer to Figure~\ref{fig:overlap} for the definition of the class overlapping settings.}
  \label{fig:cifar10_gen_ex}
\end{figure*}

\begin{figure*}[ht]
  \centering  
  \includegraphics[width=1\linewidth]{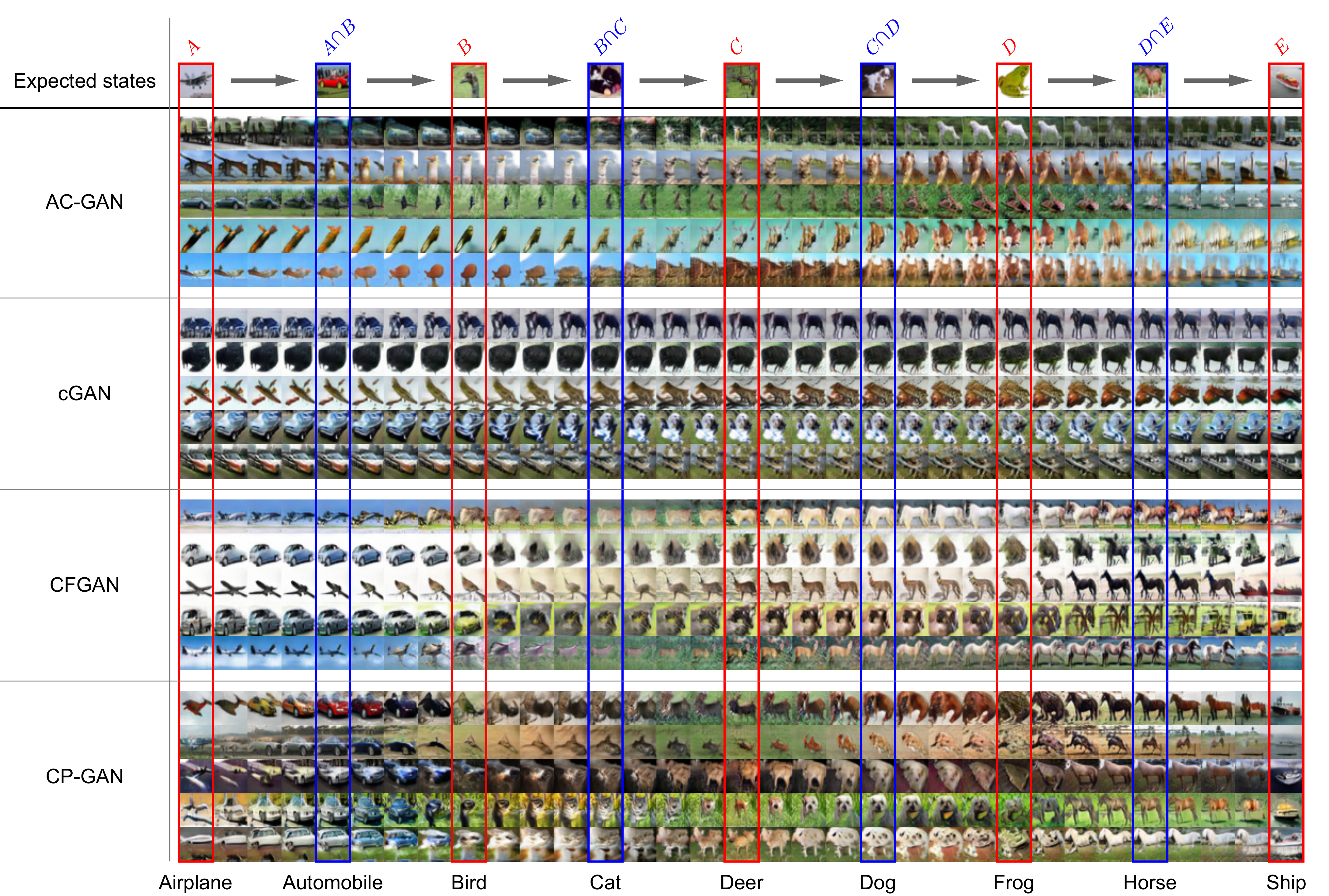}
  \caption{Continuous class-wise interpolation on CIFAR-10to5. In each row, we vary ${\bm y}^g$ continuously between classes while fixing ${\bm z}^g$. Even using the previous models (\ie, AC-GAN, cGAN, and CFGAN), it is possible to generate images continuously between classes. However, the changes are not necessarily related to the class specificity. For example, in the fifth column, $A \cap B$ (\textit{automobile}) is expected to appear, but in AC-GAN, unrelated images are generated. In contrast, CP-GAN succeeds in capturing class-distinct (surrounded by red lines) and class-mutual (surrounded by blue lines) images, and can generate them continuously on the basis of the class specificity. For example, from the first to tenth columns, CP-GAN achieves to generate $A$ (\textit{airplane}), $A \cap B$ (\textit{automobile}), and $B$ (\textit{bird}) continuously. As shown in this figure, the aim of class-distinct and class-mutual image generation is different from that of typical class-wise interpolation (or category morphing).}
  \label{fig:cifar10_interp}
\end{figure*}

\begin{figure*}[t]
  \centering
  \includegraphics[width=0.6\textwidth]{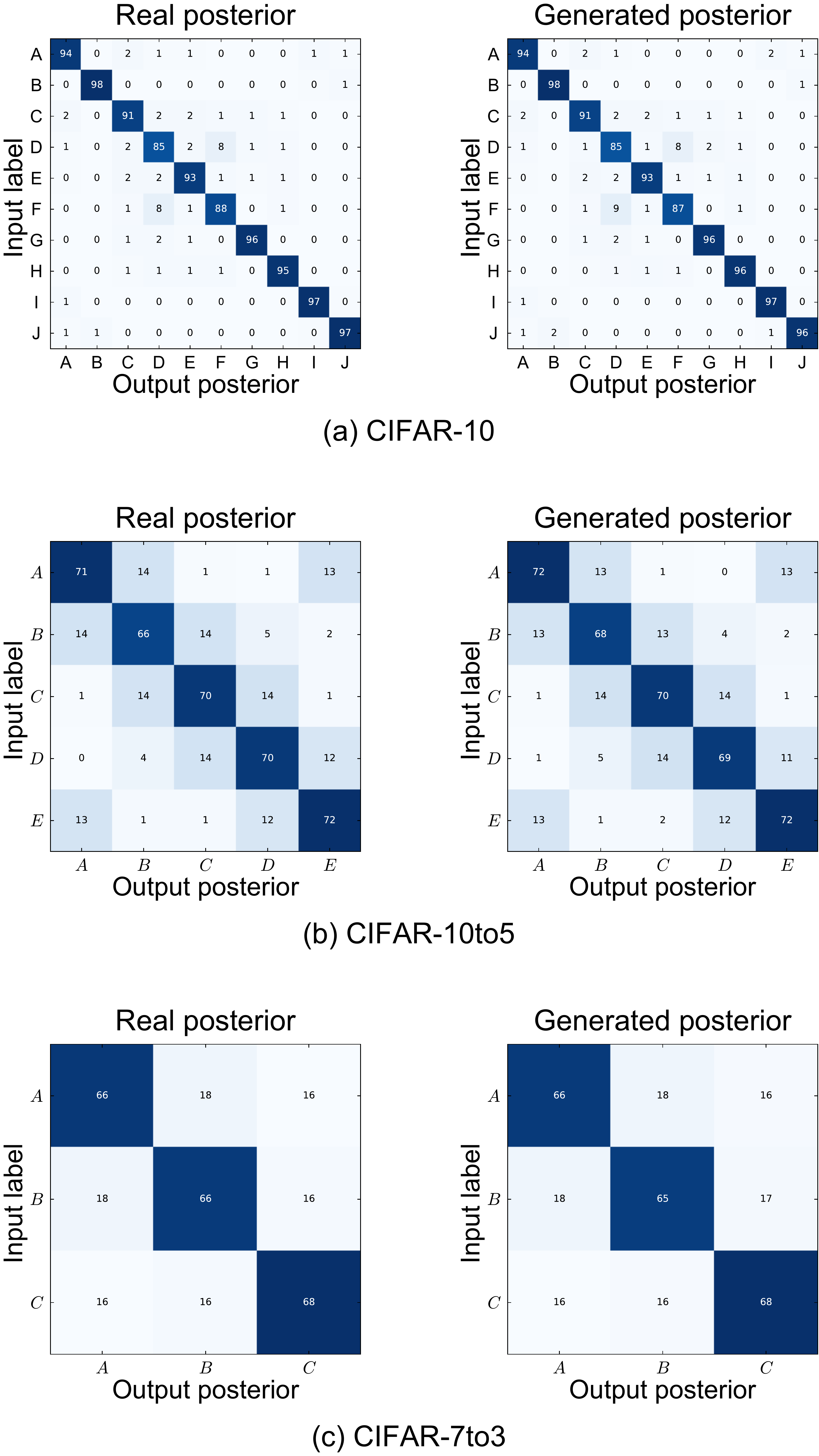}
  \caption{Comparison of the real classifier's posterior (${\bm s}^r = C({\bm y} | {\bm x}^r)$) and generated classifier's posterior using pGAN (${\bm s}^g_p = G_p({\bm z}^g_p, {\bm y}^g_p)$) in (a) CIFAR-10, (b) CIFAR-10to5, and (c) CIFAR-7to3. The values are averaging over 50$k$ samples. As shown in these figures, pGAN can generate the classifier's posterior that almost coincides with the real classifier's posterior, independently of the class-overlapping settings. Additionally, the probability distributions represent the between-class relationships (shown in Figure~\ref{fig:overlap}) reasonably well. For example, in CIFAR-10to5, the row $A$ has a higher probability in columns $B$ and $E$ in which the classes overlap. In contrast, it has a lower probability in columns $C$ and $D$ in which the classes do not overlap.}
  \label{fig:cifar10_cm}
\end{figure*}

\begin{itemize}
\item Figure~\ref{fig:cifar10_gen_ex}:
  Samples of generated images on CIFAR-10, CIFAR-10to5, and CIFAR-7to3. This is the extended version of Figure~\ref{fig:cifar10_gen}.
\item Figure~\ref{fig:cifar10_interp}:
  Continuous class-wise interpolation on CIFAR-10to5
\item Figure~\ref{fig:cifar10_cm}:
  Comparison of the real classifier's posterior and generated classifier's posterior using pGAN
\end{itemize}

\subsection{Extended results of Section~\ref{subsec:exp_real}}
\label{subsec:ex_clothing1m}

\begin{figure*}[ht]
  \centering  
  \includegraphics[width=1\linewidth]{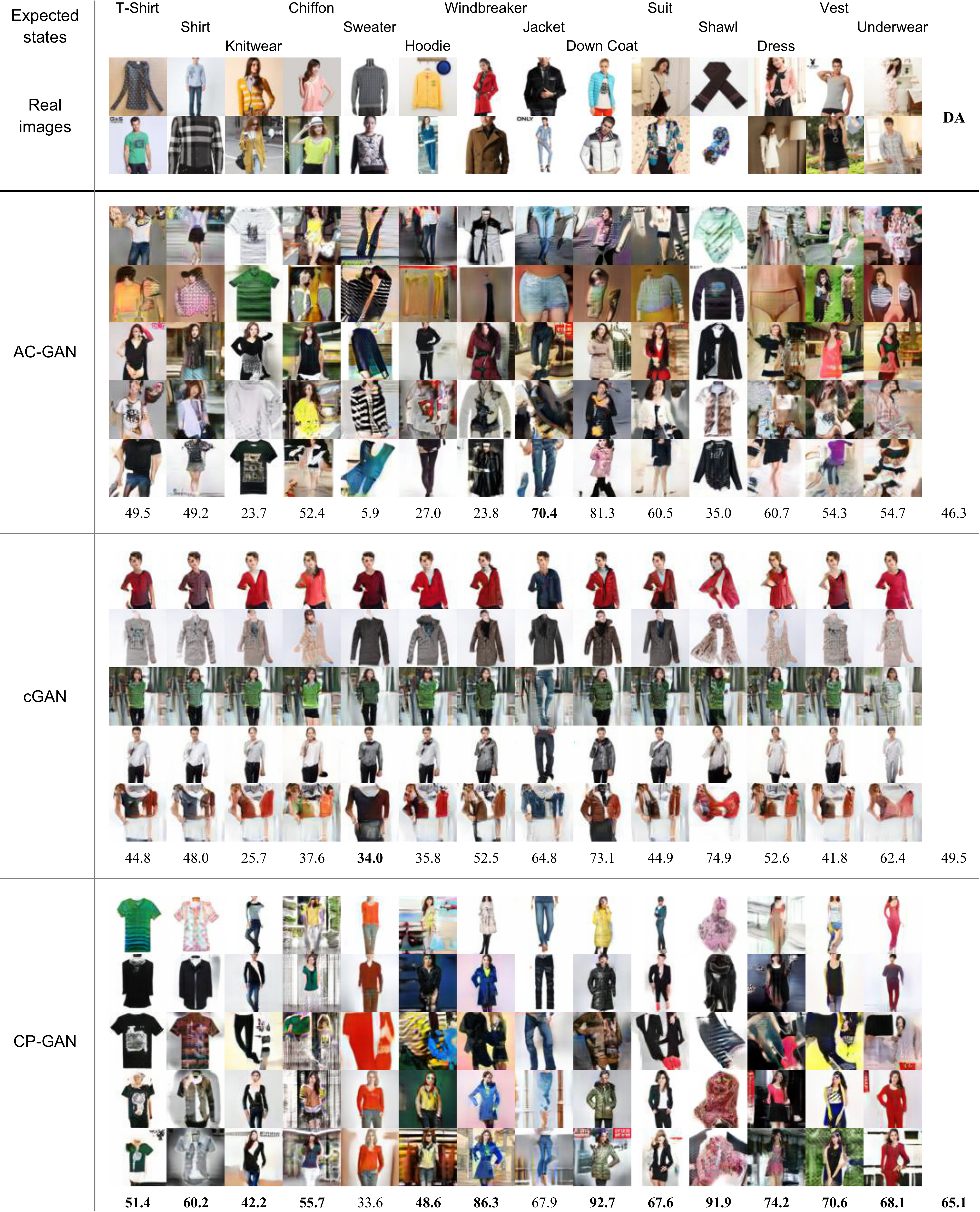}
  \caption{Samples of generated images on Clothing1M for class-distinct states (\ie, ${\bm y}^g$ is a one-hot vector). This is the extended version of Figure~\ref{fig:clothing1m_gen}. Each row contains samples generated from a fixed ${\bm z}^g$ and a varied ${\bm y}^g$. Each column includes samples generated from the same ${\bm y}^g$. Number below images indicates the per-class DA. The scores confirm that CP-GAN can generate the class-distinct (\ie, more classifiable) images by selectively using the class-distinct states (\ie, ${\bm y}^g$ is a one-hot vector).}
  \label{fig:clothing1m_gen_ex}
\end{figure*}

\begin{figure*}[t]
  \centering
  \includegraphics[width=1\textwidth]{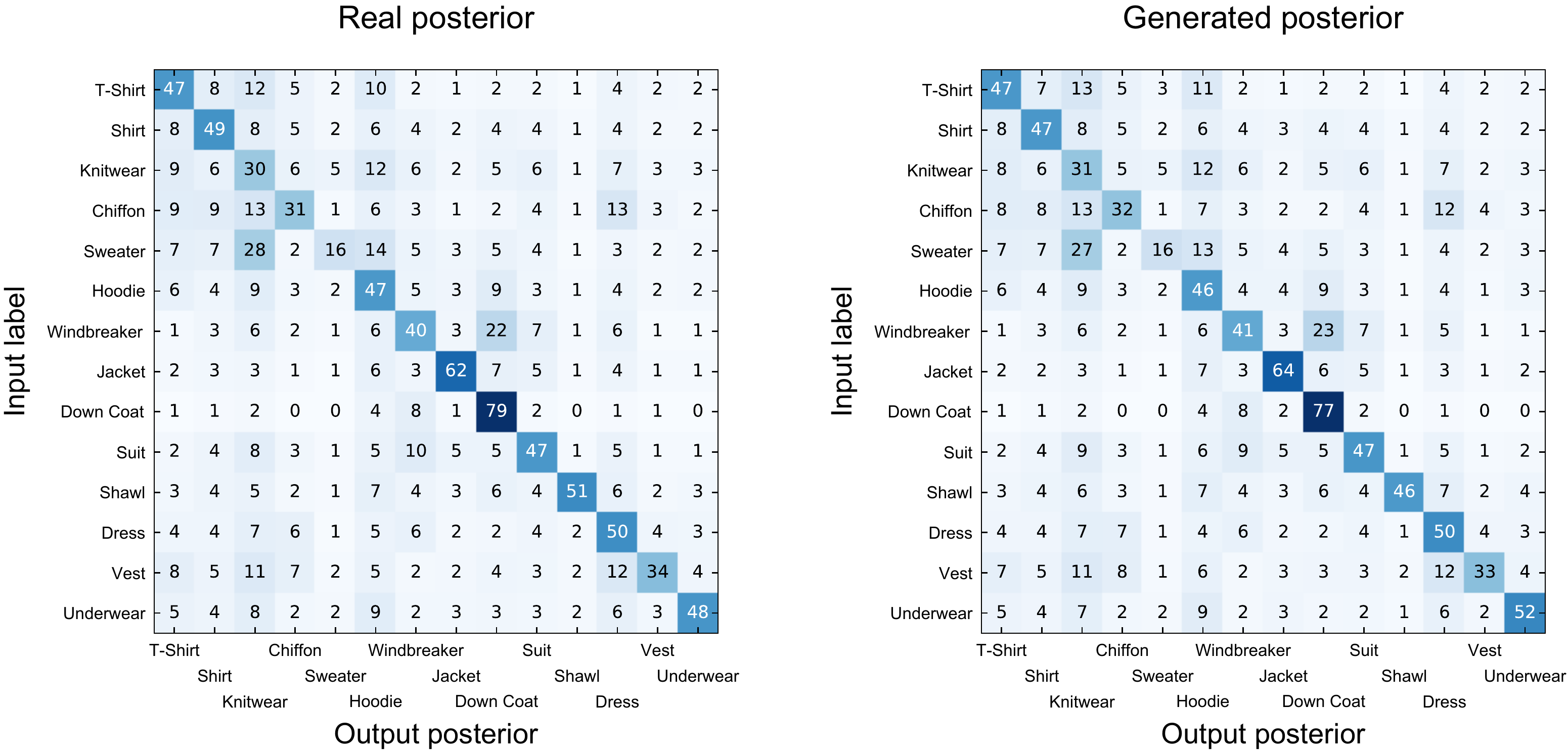}
  \caption{Comparison of the real classifier's posterior (${\bm s}^r = C({\bm y} | {\bm x}^r)$) and generated classifier's posterior using pGAN (${\bm s}^g_p = G_p({\bm z}^g_p, {\bm y}^g_p)$) in Clothing1M. The values are averaging over 50$k$ samples. These figures confirm that pGAN can generate the classifier's posterior that coincides with the real classifier's posterior also in the real-world class-overlapping data.}
  \label{fig:clothing1m_cm}
\end{figure*}

\begin{itemize}
\item Figure~\ref{fig:clothing1m_gen_ex}:
  Samples of generated images on Clothing1M. This is the extended version of Figure~\ref{fig:clothing1m_gen}.
\item Figure~\ref{fig:clothing1m_cm}:
  Comparison of the real classifier's posterior and generated classifier's posterior using pGAN
\end{itemize}

\subsection{Extended results of Appendix~\ref{sec:exp_pix2pix}}
\label{subsec:ex_pix2pix}

\begin{figure*}[ht]
  \centering  
  \includegraphics[width=0.95\linewidth]{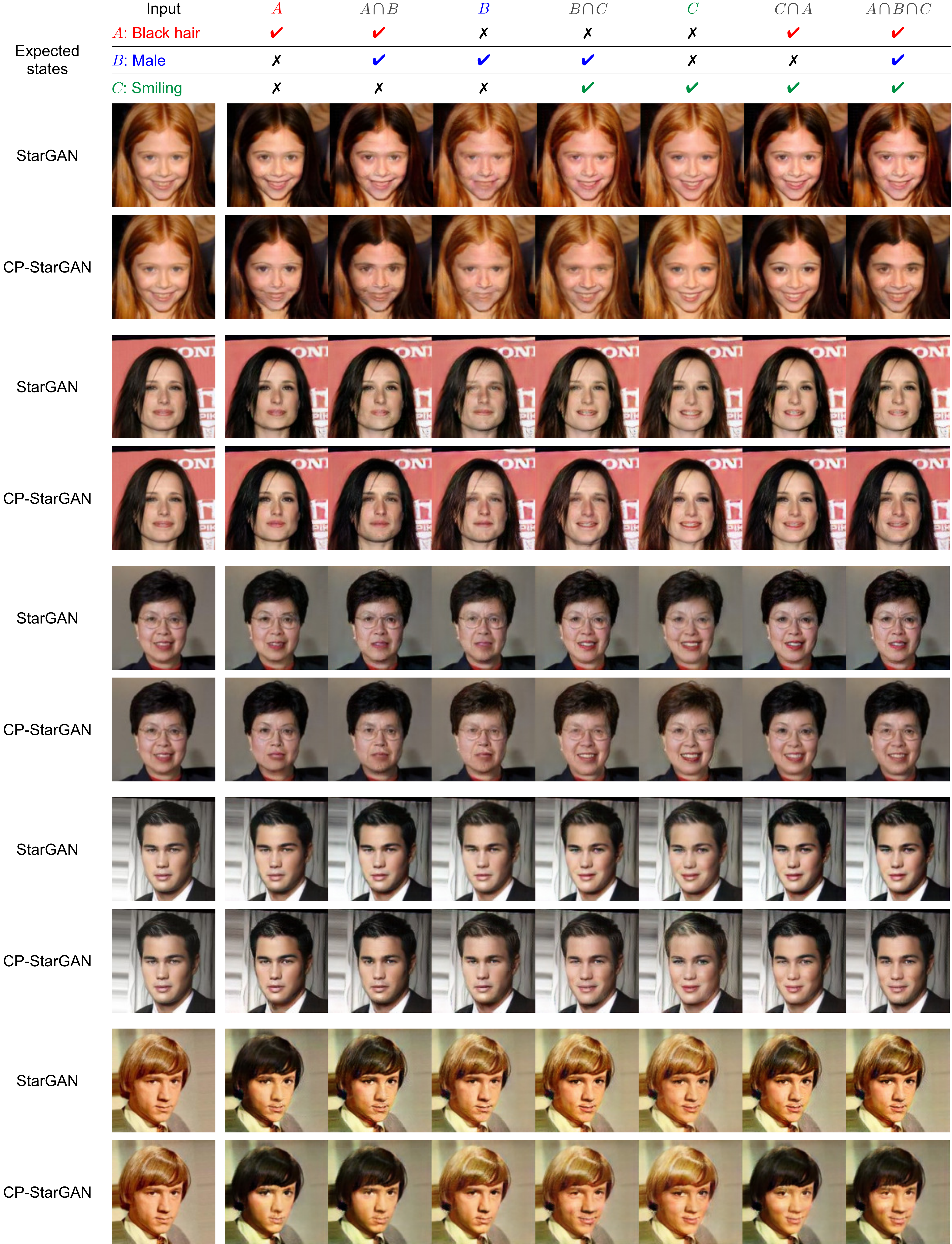}
  \caption{Samples of translated images on CelebA. This is the extended version of Figure~\ref{fig:pix2pix_gen}. Note that in training, we only know the dataset identifiers indicated by colorful font (\ie, $A$ (\textit{black hair}), $B$ (\textit{male}), and $C$ (\textit{smiling})) and class-distinct and class-mutual representations (\eg, $A \cap \neg B \cap \neg C$ (\textit{black hair}, \textit{not male}, and \textit{not smiling})) are found through learning.}
  \label{fig:pix2pix_gen_ex}
\end{figure*}

\begin{itemize}
\item Figure~\ref{fig:pix2pix_gen_ex}:
  Samples of translated images on CelebA. This is the extended version of Figure~\ref{fig:pix2pix_gen}.
\end{itemize}

\subsection{Additional results on MNIST}
\label{subsec:add_mnist}

\begin{figure*}[hb]
  \centering  
  \includegraphics[width=\linewidth]{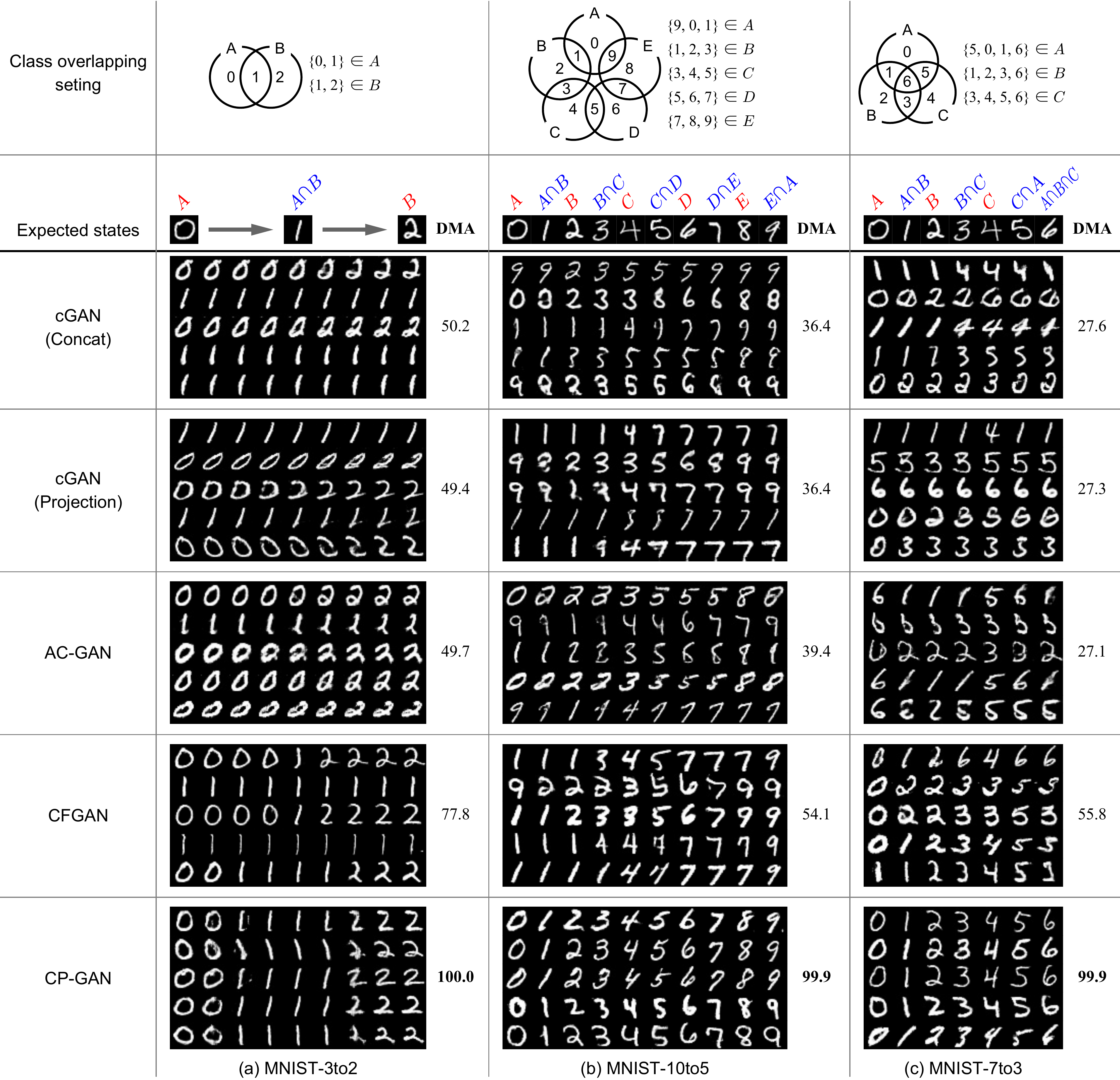}
  \caption{Samples of generated images on MNIST in class overlapping settings. We consider the situation in which each digit is divided into each class, as shown in the first row. For example, in MNIST-10to5 (b), class $A$ contains digits $0$, $1$, and $9$, and class $B$ contains $1$, $2$, and $3$. During the training, we only used class labels denoted by alphabet letters ($A, B, \dots$) as supervision. We did not use class labels of digits ($0, 1, \dots$) as supervision. We had to derive them through learning. Each column contains samples generated from the same ${\bm y}^g$. Each row shows generated samples with a fixed ${\bm z}^g$ and a varied ${\bm y}^g$. In particular, in (a) we varied ${\bm y}^g$ continuously between classes to highlight the difference between typical class-wise interpolation (or category morphing) and class-distinct and class-mutual image generation. Although in the former it is possible to generate images continuously between classes by varying ${\bm y}^g$, the changes are not necessarily related to the class specificity. In contrast, CP-GAN succeeds in selectively generating class-distinct (red font) and class-mutual (blue font) images even when the between-class relationships are not given as supervision.}
  \label{fig:mnist_gen_ex}
\end{figure*}

\begin{itemize}
\item Figure~\ref{fig:mnist_gen_ex}:
  Samples of generated images on MNIST-3to2, MNIST-10to5, and MNIST-7to3
\end{itemize}

\clearpage
\section{Details of experimental setup}
\label{sec:details}

In this appendix, we describe the details of the network architectures and training settings for each dataset.

\smallskip\noindent\textbf{Notation.}
In the description of network architectures, we use the following notation.
\begin{itemize}
  \setlength{\parskip}{1pt}
  \setlength{\itemsep}{1pt}
\item FC:
  Fully connected layer
\item Conv:
  Convolutional layer
\item Deconv:
  Deconvolutional (\ie, fractionally strided convolutional) layer
\item BN:
  Batch normalization~\cite{SIoffeICML2015}
\item IN:
  Instance normalization~\cite{DUlyanovArXiv2016}
\item ReLU:
  Rectified unit~\cite{VNairICML2010}
\item LReLU:
  Leaky rectified unit~\cite{AMaasICML2013,BXuICMLW2015}
\item ResBlock:
  Residual block~\cite{KHeCVPR2016}
\end{itemize}

In the description of training settings for GANs,
we use the following notation. Note that we used the Adam optimizer~\cite{DPKingmaICLR2015} for all GAN training.
\begin{itemize}
  \setlength{\parskip}{1pt}
  \setlength{\itemsep}{1pt}
\item $\alpha$:
  Learning rate of Adam
\item $\beta_{1}$:
  The first order momentum parameter of Adam
\item $\beta_{2}$:
  The second order momentum parameter of Adam
\item $n_{D}$:
  The number of updates of $D$ per one update of $G$
\end{itemize}

\subsection{Toy example}
\label{subsec:detail_toy}

\medskip\noindent\textbf{Generative model.}
The generative model for the toy example (\ie, two-class Gaussian distributions), which was used for the experiments discussed in Section~\ref{subsec:theory}, is shown in Table~\ref{tab:arch_toy}. As a GAN objective, we used WGAN-GP~\cite{IGulrajaniNIPS2017} and trained the model using the Adam optimizer~\cite{DPKingmaICLR2015} with a minibatch of size 256. We set the parameters to the default values of WGAN-GP for toy datasets,\footnote{We refer to the source code provided by the authors of WGAN-GP: \url{https://github.com/igul222/improved_wgan_training}.} \ie, $\lambda_{\rm GP} = 0.1$, $n_D = 5$, $\alpha = 0.0001$, $\beta_1 = 0.5$, and $\beta_2 = 0.9$. We set $\lambda^r = 1$ and $\lambda^g = 1$. We trained $G$ for $100k$ iterations, letting $\alpha$ linearly decay to 0 over $100k$ iterations.

\begin{table}[ht]
  \centering
  \begin{tabular}{cc}
    \begin{minipage}[t]{0.4\textwidth}
      \centering
      \scalebox{0.75}{
        \begin{tabular}{c} \bhline{1pt}
          {\bf Generator} $G({\bm z}^g, {\bm y}^g)$ \\ \bhline{0.75pt}
          ${\bm z}^g \in \mathbb{R}^2 \sim {\cal N}(0, I)$ $+$ ${\bm y}^g$ \\ \hline
          FC $\rightarrow$ 512, ReLU \\ \hline
          FC $\rightarrow$ 512, ReLU \\ \hline
          FC $\rightarrow$ 512, ReLU \\ \hline
          FC $\rightarrow$ 2
          \\ \bhline{1pt}
        \end{tabular}
      }
    \end{minipage}

    \hfill{\hspace{0.02\textwidth}}
    
    \begin{minipage}[t]{0.58\textwidth}
      \centering
      \scalebox{0.75}{
        \begin{tabular}{c} \bhline{1pt}
          {\bf Discriminator} $D({\bm x})$ /
          {\bf Auxiliary classifier} $C({\bm y}|{\bm x})$ \\ \bhline{0.75pt}
          ${\bm x} \in \mathbb{R}^2$ \\ \hline
          FC $\rightarrow$ 512, ReLU \\ \hline
          FC $\rightarrow$ 512, ReLU \\ \hline
          FC $\rightarrow$ 512, ReLU \\ \hline
          FC $\rightarrow$ 1 for $D$,
          FC $\rightarrow$ 2 for $C$
          \\ \bhline{1pt}
        \end{tabular}
      }
    \end{minipage}
  \end{tabular}
  \vspace{2mm}
  \caption{Generative model for toy example.}
  \label{tab:arch_toy}
\end{table}

\subsection{CIFAR-10}
\label{subsec:details_cifar10}

\medskip\noindent\textbf{Generative model for CP-GAN.}
The generative model for CIFAR-10, which was used for the experiments discussed in Section~\ref{subsec:exp_controlled}, is shown in Table~\ref{tab:arch_cifar10}. We tested two networks: WGAN-GP ResNet~\cite{IGulrajaniNIPS2017}\footnote{We refer to the source code provided by the authors of WGAN-GP: \url{https://github.com/igul222/improved_wgan_training}.} and SN-GAN ResNet~\cite{TMiyatoICLR2018b}.\footnote{We refer to the source code provided by the authors of SN-GAN: \url{https://github.com/pfnet-research/sngan_projection}.} Between these networks, the basic components are the same but the feature size in $G$ (128 in WGAN-GP ResNet but 256 in SN-GAN ResNet) and the global pooling method in $D$ (global mean pooling is used in WGAN-GP ResNet but global sum pooling is used in SN-GAN ResNet) are different. Regarding the dropout position, we also refer to CT-GAN ResNet~\cite{XWeiICLR2018}.\footnote{We refer to the source code provided by the authors of CT-GAN: \url{https://github.com/biuyq/CT-GAN/blob/master/CT-GANs}.} We used conditional batch normalization (CBN)~\cite{VDumoulinICLR2017b,HdVriesNIPS2017} to make $G$ conditioned on ${\bm y}^g$. In CP-GAN, to represent class-mixture states, we combined the CBN parameters (\ie, scale and bias parameters) with the weights calculated by the classifier's posterior. We trained the model using the Adam optimizer~\cite{DPKingmaICLR2015} with a minibatch of size 64 and 128 for $D$ and $G$, respectively. We set the parameters to the default values of WGAN-GP ResNet and SN-GAN ResNet, \ie, $n_D = 5$, $\alpha = 0.0001$, $\beta_1 = 0$, and $\beta_2 = 0.9$. We set $\lambda_{\rm GP} = 10$ for WGAN-GP ResNet. We set $\lambda^r = 1$. With regard to $\lambda^g$, we compared the performance when alternating the value of the parameter in Table~\ref{tab:eval_cifar10}. Following~\cite{IGulrajaniNIPS2017}, in WGAN-GP ResNet, we trained $G$ for $100k$ iterations, letting $\alpha$ linearly decay to 0 over $100k$ iterations. Following~\cite{TMiyatoICLR2018b}, in SN-GAN ResNet, we trained $G$ for $50k$ iterations, letting $\alpha$ linearly decay to 0 over $50k$ iterations.

\begin{table}[ht]
  \centering
  \begin{tabular}{cc}
    \begin{minipage}[t]{0.4\textwidth}
      \centering
      \scalebox{0.75}{
        \begin{tabular}{c}
          \bhline{1pt}
          {\bf Generator} $G({\bm z}^g, {\bm y}^g)$ \\ \bhline{0.75pt}
          ${\bm z}^g \in \mathbb{R}^{128} \sim {\cal N}(0, I)$,
          ${\bm y}^g$ \\ \hline
          FC $\rightarrow$ $4 \times 4 \times ch$ \\ \hline
          ResBlock up $ch$
          \\ \hline
          ResBlock up $ch$
          \\ \hline
          ResBlock up $ch$
          \\ \hline
          BN, ReLU
          \\ \hline
          $3 \times 3$ stride=1 Conv 3, Tanh
          \\ \bhline{1pt}
        \end{tabular}
      }
    \end{minipage}

    \hfill{\hspace{0.02\textwidth}}
    
    \begin{minipage}[t]{0.58\textwidth}
      \centering
      \scalebox{0.75}{
        \begin{tabular}{c} \bhline{1pt}
          {\bf Discriminator} $D({\bm x})$ /
          {\bf Auxiliary classifier} $C({\bm y} | {\bm x})$ \\ \bhline{0.75pt}
          RGB image ${\bm x} \in \mathbb{R}^{32 \times 32 \times 3}$ \\ \hline
          ResBlock down 128
          \\ \hline
          ResBlock down 128 \\
          0.2 Dropout
          \\ \hline
          ResBlock down 128 \\
          0.5 Dropout
          \\ \hline
          ResBlock 128 \\
          0.5 Dropout
          \\ \hline
          ReLU
          \\ \hline
          Global pooling
          \\ \hline
          FC $\rightarrow 1$ for $D$,
          FC $\rightarrow c$ for $C$
          \\ \bhline{1pt}
        \end{tabular}
      }
    \end{minipage}
  \end{tabular}
  \vspace{2mm}
  \caption{Generative model for CIFAR-10. In WGAN-GP ResNet, we set $ch = 128$ in $G$ and used global mean pooling in $D$. In SN-GAN ResNet, we set $ch = 256$ in $G$ and used global sum pooling in $D$.}
  \label{tab:arch_cifar10}
\end{table}

\medskip\noindent\textbf{Generative model for pGAN.}
The generative model used for pGAN, is shown in Table~\ref{tab:arch_cp}. We used cGAN with the \textit{concat.} discriminator~\cite{MMirzaArXiv2014}. The dimension of ${\bm z}^g_p$ is the same as that of ${\bm y}^g_p$. As a GAN objective, we used WGAN-GP~\cite{IGulrajaniNIPS2017} and trained the model using the Adam optimizer~\cite{DPKingmaICLR2015} with a minibatch of size 256. We set the parameters to the default values of WGAN-GP for toy datasets,\footnote{We refer to the source code provided by the authors of WGAN-GP: \url{https://github.com/igul222/improved_wgan_training}.} \ie, $\lambda_{\rm GP} = 0.1$, $n_D = 5$, $\alpha = 0.0001$, $\beta_1 = 0.5$, and $\beta_2 = 0.9$. We trained $G_p$ for $100k$ iterations, letting $\alpha$ linearly decay to 0 over $100k$ iterations.

\begin{table}[ht]
  \centering
  \begin{tabular}{cc}
    \begin{minipage}[t]{0.4\textwidth}
      \centering
      \scalebox{0.75}{
        \begin{tabular}{c} \bhline{1pt}
          {\bf Generator} $G_p({\bm z}^g_p, {\bm y}^g_p)$ \\ \bhline{0.75pt}
          ${\bm z}_p \in \mathbb{R}^{c} \sim {\cal N}(0, I)$, ${\bm y}^g_p$ \\ \hline
          FC $\rightarrow$ 512, ReLU \\ \hline
          FC $\rightarrow$ 512, ReLU \\ \hline
          FC $\rightarrow$ 512, ReLU \\ \hline
          FC $\rightarrow$ $c$
          \\ \bhline{1pt}
        \end{tabular}
      }
    \end{minipage}

    \hfill{\hspace{0.02\textwidth}}
    
    \begin{minipage}[t]{0.58\textwidth}
      \centering
      \scalebox{0.75}{
        \begin{tabular}{c} \bhline{1pt}
          {\bf Discriminator} $D_p({\bm s}, {\bm y})$ \\ \bhline{0.75pt}
          ${\bm s} \in \mathbb{R}^c$, ${\bm y}$ \\ \hline
          FC $\rightarrow$ 512, ReLU + ${\bm y}$ \\ \hline
          FC $\rightarrow$ 512, ReLU + ${\bm y}$ \\ \hline
          FC $\rightarrow$ 512, ReLU + ${\bm y}$ \\ \hline
          FC $\rightarrow$ 1 for $D$
          \\ \bhline{1pt}
        \end{tabular}
      }
    \end{minipage}
  \end{tabular}
  \vspace{2mm}
  \caption{Generative model for pGAN.}
  \label{tab:arch_cp}
\end{table}

\medskip\noindent\textbf{Classifier model used for evaluation.}
The classifier model used for the CIFAR-10 DMA evaluation is shown in Table~\ref{tab:arch_cifar10_cls}. We trained the model using the SGD optimizer with a minibatch of size 128. We set the initial learning rate to 0.1 and divided it by 10 when the iterations are $40k$, $60k$, and $80k$. We trained the model for $100k$ iterations in total. The accuracy score for the CIFAR-10 test data was $94.7\%$.

\begin{table}[ht]  
  \centering
  \scalebox{0.75}{
    \begin{tabular}{c} \bhline{1pt}
      {\bf Classifier}
      \\ \bhline{0.75pt}
      RGB image ${\bm x} \in \mathbb{R}^{32 \times 32 \times 3}$ \\ \hline
      $3 \times 3$ stride=1 Conv 128, BN, ReLU \\
      $3 \times 3$ stride=1 Conv 128, BN, ReLU \\
      MaxPool \\ \hline

      $3 \times 3$ stride=1 Conv 256, BN, ReLU \\
      $3 \times 3$ stride=1 Conv 256, BN, ReLU \\
      MaxPool \\ \hline

      $3 \times 3$ stride=1 Conv 512, BN, ReLU \\
      $3 \times 3$ stride=1 Conv 512, BN, ReLU \\
      $3 \times 3$ stride=1 Conv 512, BN, ReLU \\
      $3 \times 3$ stride=1 Conv 256, BN, ReLU \\
      MaxPool \\ \hline

      FC $\rightarrow$ 1024, ReLU \\
      0.5 Dropout \\ \hline
      FC $\rightarrow$ 1024, ReLU \\
      0.5 Dropout \\ \hline
      FC $\rightarrow$ $c$
      \\ \bhline{1pt}
    \end{tabular}
  }
  \vspace{2mm}
  \caption{Classifier model used for CIFAR-10 DMA evaluation.}
  \label{tab:arch_cifar10_cls}
\end{table}

\subsection{Clothing1M}
\label{subsec:details_clothing1M}

\medskip\noindent\textbf{Generative model.}
The generative model for Clothing1M, which was used for the experiments discussed in Section~\ref{subsec:exp_real}, is shown in Table~\ref{tab:arch_clothing1m}. As GAN configurations, we used WGAN-GP ResNet~\cite{IGulrajaniNIPS2017} for AC-GAN and CP-GAN and used SN-GAN ResNet~\cite{TMiyatoICLR2018b} for cGAN with projection discriminator. We trained the model using the Adam optimizer~\cite{DPKingmaICLR2015} with a minibatch of size 66. We set the parameters to the default values of WGAN-GP ResNet for $64 \times 64$ images,\footnote{We refer to the source code provided by the authors of WGAN-GP: \url{https://github.com/igul222/improved_wgan_training}.} \ie, $n_D = 5$, $\alpha = 0.0001$, $\beta_1 = 0$, and $\beta_2 = 0.9$. We set $\lambda_{\rm GP} = 10$ for WGAN-GP ResNet. We set the trade-off parameters for the auxiliary classifier to $\lambda^r = 1$ and $\lambda^g = 0.1$. We trained $G$ for $200k$ iterations.

\begin{table}[ht]
  \centering
  \begin{tabular}{cc}
    \begin{minipage}[t]{0.4\textwidth}
      \centering
      \scalebox{0.75}{
        \begin{tabular}{c}
          \bhline{1pt}
          {\bf Generator} $G({\bm z}^g, {\bm y}^g)$ \\ \bhline{0.75pt}
          ${\bm z}^g \in \mathbb{R}^{128} \sim {\cal N}(0, I)$,
          ${\bm y}^g$ \\ \hline
          FC $\rightarrow$ $4 \times 4 \times 512$ \\ \hline
          ResBlock up 512
          \\ \hline
          ResBlock up 256
          \\ \hline
          ResBlock up 128
          \\ \hline
          ResBlock up 64
          \\ \hline
          BN, ReLU
          \\ \hline
          $3 \times 3$ stride=1 Conv 3, Tanh
          \\ \bhline{1pt}
        \end{tabular}
      }
    \end{minipage}

    \hfill{\hspace{0.02\textwidth}}
    
    \begin{minipage}[t]{0.58\textwidth}
      \centering
      \scalebox{0.75}{
        \begin{tabular}{c} \bhline{1pt}
          {\bf Discriminator} $D({\bm x})$ /
          {\bf Auxiliary classifier} $C({\bm y} | {\bm x})$ \\ \bhline{0.75pt}
          RGB image ${\bm x} \in \mathbb{R}^{64 \times 64 \times 3}$ \\ \hline
          $3 \times 3$ stride=1 Conv 64
          \\ \hline
          ResBlock down 128 \\
          0.2 Dropout
          \\ \hline
          ResBlock down 256 \\
          0.2 Dropout
          \\ \hline
          ResBlock down 512 \\
          0.5 Dropout
          \\ \hline
          ResBlock down 512 \\
          0.5 Dropout
          \\ \hline
          FC $\rightarrow 1$ for $D$,
          FC $\rightarrow c$ for $C$
          \\ \bhline{1pt}          
        \end{tabular}
      }
    \end{minipage}
  \end{tabular}
  \vspace{2mm}
  \caption{Generative model for Clothing1M.}
  \label{tab:arch_clothing1m}
\end{table}

\medskip\noindent\textbf{Classifier model used for evaluation.}
The classifier model used for the Clothing1M DA evaluation is shown in Table~\ref{tab:arch_clothing1m_cls}. We trained the model using the SGD optimizer with a minibatch of size 129. We set the initial learning rate to 0.1 and divided it by 10 when the iterations are $40k$, $60k$, and $80k$. We trained the model for $100k$ iterations in total. The accuracy score for clean labeled data was $71.8\%$.

\begin{table}[ht]  
  \centering
  \scalebox{0.75}{
    \begin{tabular}{c} \bhline{1pt}
      {\bf Classifier}
      \\ \bhline{0.75pt}
      RGB image ${\bm x} \in \mathbb{R}^{64 \times 64 \times 3}$ \\ \hline
      $3 \times 3$ stride=1 Conv 64, BN, ReLU \\
      $4 \times 4$ stride=2 Conv 64, BN, ReLU \\
      0.5 Dropout \\ \hline
      
      $3 \times 3$ stride=1 Conv 128, BN, ReLU \\
      $3 \times 3$ stride=1 Conv 128, BN, ReLU \\
      $4 \times 4$ stride=2 Conv 128, BN, ReLU \\
      0.5 Dropout \\ \hline

      $3 \times 3$ stride=1 Conv 256, BN, ReLU \\
      $3 \times 3$ stride=1 Conv 256, BN, ReLU \\
      $4 \times 4$ stride=2 Conv 256, BN, ReLU \\
      0.5 Dropout \\ \hline

      $3 \times 3$ stride=1 Conv 512, BN, ReLU \\
      $3 \times 3$ stride=1 Conv 512, BN, ReLU \\
      $3 \times 3$ stride=2 Conv 256, BN, ReLU \\
      0.5 Dropout \\ \hline

      $3 \times 3$ stride=1 Conv 512, BN, ReLU \\
      $3 \times 3$ stride=1 Conv 512, BN, ReLU \\
      $3 \times 3$ stride=1 Conv 256, BN, ReLU \\
      0.5 Dropout \\ \hline
      
      FC $\rightarrow$ $c$
      \\ \bhline{1pt}
    \end{tabular}
  }
  \vspace{2mm}
  \caption{Classifier model used for Clothing1M DA evaluation.}
  \label{tab:arch_clothing1m_cls}
\end{table}

\subsection{CelebA (image-to-image translation)}
\label{subsec:details_pix2pix}

\medskip\noindent\textbf{Generative model.}
The generative model for CelebA on image-to-image-translation tasks, which was used for the experiments discussed in Appendix~\ref{sec:exp_pix2pix}, is shown in Table~\ref{tab:arch_pix2pix}. The network architecture is the same as that of StarGAN~\cite{YChoiCVPR2018}.\footnote{We refer to the source code provided by the authors of StarGAN: \url{https://github.com/yunjey/StarGAN}.} As a GAN objective, we used WGAN-GP~\cite{IGulrajaniNIPS2017} and trained the model using the Adam optimizer~\cite{DPKingmaICLR2015} with a minibatch of size 16. We set the parameters to the default values of the StarGAN, \ie, $\lambda_{\rm rec} = 10$, $\lambda_{\rm GP} = 10$, $n_D = 5$, $\alpha = 0.0001$, $\beta_1 = 0.5$, and $\beta_2 = 0.999$. We set the trade-off parameters for the auxiliary classifier to $\lambda^r = 1$ and $\lambda^g = 1$. We trained the $G$ for $50k$ iterations and let $\alpha$ linearly decay to 0 over the last $25k$ iterations.

\begin{table}[ht]
  \centering
  \begin{tabular}{cc}
    \begin{minipage}[t]{0.4\textwidth}
      \centering
      \scalebox{0.75}{
        \begin{tabular}{c} \bhline{1pt}
          {\bf Generator} $G({\bm x}^r, {\bm y}^g)$ \\ \bhline{0.75pt}
          ${\bm x}^r \in \mathbb{R}^{128 \times 128 \times 3}$,
          ${\bm y}^g$ \\ \hline
          $7 \times 7$ stride=1 Conv 64, IN, ReLU \\
          $4 \times 4$ stride=2 Conv 128, IN, ReLU \\
          $4 \times 4$ stride=2 Conv 256, IN, ReLU \\ \hline
          ResBlock 256 \\
          ResBlock 256 \\
          ResBlock 256 \\
          ResBlock 256 \\
          ResBlock 256 \\
          ResBlock 256 \\ \hline
          $4 \times 4$ stride=2 Deconv 128, IN, ReLU \\
          $4 \times 4$ stride=2 Deconv  64, IN, ReLU \\
          $7 \times 7$ stride=1 Conv 3, Tanh
          \\ \bhline{1pt}
        \end{tabular}
      }
    \end{minipage}

    \hfill{\hspace{0.02\textwidth}}
    
    \begin{minipage}[t]{0.58\textwidth}
      \centering
      \scalebox{0.75}{
        \begin{tabular}{c} \bhline{1pt}
          {\bf Discriminator} $D({\bm x})$ /
          {\bf Auxiliary classifier} $C({\bm y} | {\bm x})$ \\ \bhline{0.75pt}
          RGB image ${\bm x} \in \mathbb{R}^{128 \times 128 \times 3}$ \\ \hline
          $4 \times 4$ stride=2 Conv 64, LReLU \\
          $4 \times 4$ stride=2 Conv 128, LReLU \\
          $4 \times 4$ stride=2 Conv 256, LReLU \\
          $4 \times 4$ stride=2 Conv 512, LReLU \\
          $4 \times 4$ stride=2 Conv 1024, LReLU \\
          $4 \times 4$ stride=2 Conv 2048, LReLU \\ \hline
          $3 \times 3$ stride=1 Conv 1 for $D$ \\
          $2 \times 2$ stride=1 Conv (zero pad) $c$ for $C$
          \\ \bhline{1pt}
        \end{tabular}
      }
    \end{minipage}
  \end{tabular}
  \vspace{2mm}
  \caption{Generative model for CelebA (image-to-image translation).}
  \label{tab:arch_pix2pix}
\end{table}

\medskip\noindent\textbf{Classifier model used for evaluation.}
The classifier model used for the CelebA DMA evaluation on image-to-image-translation tasks is shown in Table~\ref{tab:arch_pix2pix_cls}. We trained the model using the SGD optimizer with a minibatch of size 128. We set the initial learning rate to 0.1 and divided it by 10 when the iterations are $40k$, $60k$, and $80k$. We trained the model for $100k$ iterations in total. The accuracy scores for black hair, male, and smiling (binary classification) were $88.4\%$, $98.5\%$, and $93.0\%$, respectively.

\begin{table}[ht]  
  \centering
  \scalebox{0.75}{
    \begin{tabular}{c} \bhline{1pt}
      {\bf Classifier}
      \\ \bhline{0.75pt}
      RGB image ${\bm x} \in \mathbb{R}^{128 \times 128 \times 3}$ \\ \hline
      $3 \times 3$ stride=1 Conv 32, BN, ReLU \\
      $4 \times 4$ stride=2 Conv 32, BN, ReLU \\
      0.5 Dropout \\ \hline
      
      $3 \times 3$ stride=1 Conv 64, BN, ReLU \\
      $4 \times 4$ stride=2 Conv 64, BN, ReLU \\
      0.5 Dropout \\ \hline
      
      $3 \times 3$ stride=1 Conv 128, BN, ReLU \\
      $4 \times 4$ stride=2 Conv 128, BN, ReLU \\
      0.5 Dropout \\ \hline

      $3 \times 3$ stride=1 Conv 256, BN, ReLU \\
      $4 \times 4$ stride=2 Conv 256, BN, ReLU \\
      0.5 Dropout \\ \hline

      $3 \times 3$ stride=1 Conv 512, BN, ReLU \\
      $3 \times 3$ stride=1 Conv 512, BN, ReLU \\
      $3 \times 3$ stride=1 Conv 512, BN, ReLU \\
      $3 \times 3$ stride=2 Conv 256, BN, ReLU \\
      0.5 Dropout \\ \hline
      
      FC $\rightarrow$ $c$
      \\ \bhline{1pt}
    \end{tabular}
  }
  \vspace{2mm}
  \caption{Classifier model used for CelebA DMA evaluation.}
  \label{tab:arch_pix2pix_cls}
\end{table}

\subsection{MNIST}
\label{subsec:details_mnist}

\medskip\noindent\textbf{Generative model.}
The generative model for MNIST, which was used for the experiments discussed in Figure~\ref{fig:mnist_gen_ex}, is shown in Table~\ref{tab:arch_mnist}. As a GAN objective, we used the WGAN-GP~\cite{IGulrajaniNIPS2017} and train the model using the Adam optimizer~\cite{DPKingmaICLR2015} with a minibatch of size 64. We set the parameters to the default values of the WGAN-GP,\footnote{We refer to the source code provided by the authors of WGAN-GP: \url{https://github.com/igul222/improved_wgan_training}.} \ie, $\lambda_{\rm GP} = 10$, $n_D = 5$, $\alpha = 0.0001$, $\beta_1 = 0.5$, and $\beta_2 = 0.9$. We set the trade-off parameters for the auxiliary classifier to $\lambda^r = 1$ and $\lambda^g = 1$ and trained $G$ for $200k$ iterations.

\begin{table}[ht]
  \centering
  \begin{tabular}{cc}
    \begin{minipage}[t]{0.4\textwidth}
      \centering
      \scalebox{0.75}{
        \begin{tabular}{c} \bhline{1pt}
          {\bf Generator} $G({\bm z}^g, {\bm y}^g)$ \\ \bhline{0.75pt}
          ${\bm z}^g \in \mathbb{R}^{128}$, ${\bm y}^g$ \\ \hline
          FC $\rightarrow$ 4096, BN, ReLU \\
          Reshape $4 \times 4 \times 256$ \\ \hline
          $5 \times 5$ stride=2 Conv 128, BN, ReLU \\
          Cut $7 \times 7 \times 128$
          \\ \hline
          $5 \times 5$ stride=2 Conv 64, BN, ReLU
          \\ \hline
          $5 \times 5$ stride=2 Conv 1, Sigmoid
          \\ \bhline{1pt}
        \end{tabular}
      }
    \end{minipage}

    \hfill{\hspace{0.02\textwidth}}
    
    \begin{minipage}[t]{0.58\textwidth}
      \centering
      \scalebox{0.75}{
        \begin{tabular}{c} \bhline{1pt}
          {\bf Discriminator} $D({\bm x})$ /
          {\bf Auxiliary classifier} $C({\bm y}|{\bm x})$ \\ \bhline{0.75pt}
          Gray image ${\bm x} \in \mathbb{R}^{28 \times 28 \times 1}$ \\ \hline
          $5 \times 5$ stride=2 Conv 64, LReLU \\
          0.5 Dropout
          \\ \hline
          $5 \times 5$ stride=2 Conv 128, LReLU \\
          0.5 Dropout
          \\ \hline
          $5 \times 5$ stride=2 Conv 256, LReLU \\
          0.5 Dropout
          \\ \hline
          FC $\rightarrow$ 1 for $D$,
          FC $\rightarrow$ $c$ for $C$
          \\ \bhline{1pt}
        \end{tabular}
      }
    \end{minipage}
  \end{tabular}
  \vspace{2mm}
  \caption{Generative model for MNIST.}
  \label{tab:arch_mnist}
\end{table}

\medskip\noindent\textbf{Classifier model used for evaluation.}
The classifier model used for the MNIST DMA evaluation is shown in Table~\ref{tab:arch_mnist_cls}. We trained the model using the SGD optimizer with a minibatch of size 128. We set the initial learning rate to 0.1 and divided it by 10 when the iterations are $40k$, $60k$, and $80k$. We trained the model for $100k$ iterations in total. The accuracy score for the MNIST test data was $99.7\%$.

\begin{table}[ht]
  \centering
  \scalebox{0.75}{
    \begin{tabular}{c} \bhline{1pt}
      {\bf Classifier}
      \\ \bhline{0.75pt}
      Gray image ${\bm x} \in \mathbb{R}^{28 \times 28 \times 1}$ \\ \hline
      $3 \times 3$ stride=1 Conv 64, BN, ReLU \\
      $4 \times 4$ stride=2 Conv 64, BN, ReLU \\
      0.5 Dropout \\ \hline
      
      $3 \times 3$ stride=1 Conv 128, BN, ReLU \\
      $4 \times 4$ stride=2 Conv 128, BN, ReLU \\
      0.5 Dropout \\ \hline

      $3 \times 3$ stride=1 Conv 256, BN, ReLU \\
      $3 \times 3$ stride=1 Conv 256, BN, ReLU \\
      $3 \times 3$ stride=1 Conv 256, BN, ReLU \\
      $3 \times 3$ stride=2 Conv 256, BN, ReLU \\
      0.5 Dropout \\ \hline
      
      FC $\rightarrow$ $c$
      \\ \bhline{1pt}
    \end{tabular}
  }
  \vspace{2mm}
  \caption{Classifier model used for MNIST DMA evaluation.}
  \label{tab:arch_mnist_cls}
\end{table}

\end{document}